\documentclass[3p,10pt]{elsarticle}

\usepackage{graphicx}
\usepackage[caption=false]{subfig}
\usepackage{amssymb}
\usepackage{csquotes}
\usepackage{textcomp} %
\usepackage{tabularx}
\usepackage{longtable} %
\usepackage{pdflscape} %
\usepackage{hyperref}
\usepackage{multirow}

\makeatletter 
\def\ps@pprintTitle{%
  \let\@oddhead\@empty
  \let\@evenhead\@empty
  \let\@oddfoot\@empty
  \let\@evenfoot\@oddfoot
}
\makeatother 

\usepackage{fancyhdr}
\usepackage{bm}

\usepackage[nolists, tablesfirst, nomarkers]{endfloat} %

\usepackage[toc]{appendix}%

 \biboptions{super}

\begin{document}

\begin{frontmatter}

\title{Design choice and machine learning model performances}

\author{
{Rosa Arboretti\textsuperscript{a}, Riccardo Ceccato\textsuperscript{b}, Luca Pegoraro\textsuperscript{b} and Luigi Salmaso\textsuperscript{b,*}}
}
\address{\textsuperscript{a}Department of Civil, Environmental and Architectural Engineering, Università degli Studi di Padova, Padua, Italy; \textsuperscript{b}Department of Management and Engineering, Università degli Studi di Padova, Vicenza, Italy

*corresponding author: luigi.salmaso@unipd.it
}

\begin{abstract}
An increasing number of publications present the joint application of Design of Experiments (DOE) and machine learning (ML) as a methodology to collect and analyze data on a specific industrial phenomenon. However, the literature shows that the choice of the design for data collection and model for data analysis is often not driven by statistical or algorithmic advantages, thus there is a lack of studies which provide guidelines on what designs and ML models to jointly use for data collection and analysis. This article discusses the choice of design in relation to the ML model performances. A study is conducted that considers 12 experimental designs, 7 families of predictive models, 7 test functions that emulate physical processes, and 8 noise settings, both homoscedastic and heteroscedastic. The results of the research can have an immediate impact on the work of practitioners, providing guidelines for practical applications of DOE and ML.
\end{abstract}

\begin{keyword}
Simulation study \sep Physical experiments \sep Predictive analytics \sep Gaussian Process \sep Artificial Neural Networks

\end{keyword}

\end{frontmatter}

\section{Introduction}
\label{S:1}

In many industries the development of mathematical models of physical phenomena or processes is of primary interest. Objectives of such models may include prediction of one or more quantities or identification of the optimal configuration of parameters which leads to an optimization of some quantities of interest, such as a quality criterion or a cost measure. Design of Experiments (DOE) is a well-known statistical procedure that has been widely applied for decades especially for optimization, but also for the construction of surrogate models able to reliably predict one or more responses of interest. Historically, the most widespread model employed for the analysis of DOE data, which also stands at the core of the response surface methodology (RSM) \cite{MyersMontgomery2004}, is the second-order model \cite{montgomery2017design}.

Recently, with the advent of the fourth industrial revolution and the consequential increase in data availability in the industrial environment, machine learning (ML) techniques for data analysis have rapidly spread in industry. In this context, a recent trend consists in the adoption of ML algorithms for the analysis of experimental data, in lieu of the typical second-order model, as proved by the increasing number of publications dealing with these topics, as shown in \citet{arbo_peg21SLR}. However, it has also been shown that there is a lack of studies that try to provide some guidelines on what experimental designs appear to be more appropriate when ML models are selected as a tool for analysis \cite{arbo_peg21SLR}. Instead, in the vast majority of papers that jointly apply DOE and ML, the choice of one specific design appears to be driven by incidental factors such as the analyst's experience or diffusion in one specific area \cite{arbo_peg21SLR}. The aim of this article is to show in an empirical manner what designs are more promising for collection of data that is then modeled through ML techniques. As a collateral result, this will also lead to indications being provided on what predictive analytics techniques to take into consideration when analyzing DOE data. It should be pointed out here that only predictive performance of the algorithms in terms of prediction error will be taken into account, and other aspects, such as the ability to quantify uncertainty of predictions and the possibility of investigating causality, are features that should be considered when choosing an algorithm for prediction or optimization \cite{arbo_peg21SLR}. The authors encourage further research devoted to these topics, but this is outside the scope of the present work.

The paper is organized as follows: section \ref{S:2} presents the literature background to the application of ML for the analysis of DOE data; section \ref{S:3} presents the methodology and details of the simulation study conducted; section \ref{S:4} shows the results and section \ref{S:5} discusses the findings in depth. Finally, in section \ref{S:6}, a summary of the main contributions of the work and the conclusions are presented.

\section{Literature background}
\label{S:2}
In their recent article, \citet{arbo_peg21SLR} provided a comprehensive overview of the topics of DOE and ML applied in a product innovation (PI) setting. \citet{arbo_peg21SLR} highlight both the advantages and challenges of the application of ML in PI, discuss the implications of a joint adoption of DOE and ML, and identify the most common experimental designs employed and ML algorithms chosen for analysis. From their work, some clear advantages of the adoption of ML for data analysis appear evident, including the ability of ML to appropriately model data without requiring assumptions on the underlying distribution, the ability to model complex non-linear relationships, and the general capability of providing a better fit to the data, thus ensuring more accurate predictions and the requirement for a less strict design structure, to the point that undesigned data can be effectively modeled. Additionally, a proper adoption of a DOE+ML framework can enhance further advantages, including the minimization of the number of trials through the identification of the most informative combinations of the factor levels, the ability to explore experimental regions omitted by DOE alone, the possibility to automate the experimentation procedure, and others \cite{arbo_peg21SLR}.

An in-depth analysis of the type of DOE used in combination with ML modeling was also conducted \cite{arbo_peg21SLR}, revealing high fragmentation in the choice of designs. The most recurring class of DOEs comes from the RSM literature including Central Composite Designs \cite{CCD} (CCDs), Box-Behnken Designs \cite{BBD} (BBDs), and full and fractional factorial designs (FFDs). Another relevant group of experimental designs comes from the Robust Parameter Design (RPD) literature \cite{taguchi1979introduction, taguchi1986introduction, taguchi1987system}, while space-filling designs are seldomly applied \cite{Joseph2016}. Overall, the non-sequential experimentation procedure is still predominant \cite{arbo_peg21SLR}, despite the fact it has been shown that several advantages are embedded in a sequential strategy \cite{box1999statistics, vining2011, jensen2018open}. The overall picture in regards to the choice of design in a DOE+ML framework is not that surprising as another finding is that physical experiments are by far more common than computer experiments \cite{arbo_peg21SLR}, meaning that traditional designs from RSM are favored. The main reason is that such designs have been specifically developed for the optimization of physical systems and they can count on a very well established literature set that proposes solutions for many of the typical problems which characterize the industrial context, e.g. restrictions to randomization with the split-plot designs \cite{cortes2018response}, or the need for an estimation of many effects while minimizing the experimental effort with definitive screening designs \cite{jones2011class}. Other designs, such as space-filling designs, have been developed for computer experiments in which the concepts of randomization and replication, that are at the base of traditional DOE, no longer count because experiments are typically deterministic and the frequent change in factor levels is possible and does not imply a cost. Also, the preference for non-sequential experimentation is likely a consequence of physical experimentation, since companies may adopt strict procedures and protocols that discourage the adoption of a sequential approach. For instance, access to a system under investigation may be limited because experimentation can cause a delay to production deadlines and management may be unwilling to reserve the required resources for a prolonged period. Thus, in many situations, execution of an entire DOE in one single batch of trials is preferable from a production planning perspective.

\citet{arbo_peg21SLR} also provide a picture of the ML models most widely applied to DOE data for product innovation, revealing extensive use of Artificial Neural Networks (ANNs), and sporadic adoption of other prediction algorithms, such as Gaussian Processes (GPs) and Support Vector Regression models (SVMs).

\section{Design and model choice}
\label{S:3}
In this section we detail the procedure and simulation study conducted to assess the performances of different designs paired with different ML models for data analysis. In section \ref{S:3.1} we present the research objective and its boundaries, in section \ref{S:3.2} and \ref{S:3.3} we briefly review the chosen designs and the ML algorithms respectively. In sections \ref{S:3.4} and \ref{S:3.5} we describe the selected test functions and different noise structures and levels, and in section \ref{S:3.6} we detail the ranking procedure adopted to obtain the final rank of designs and algorithms.

\subsection{Objective and boundaries}
\label{S:3.1}
The objective of the simulation study is to provide, for the first time, an empirical assessment of the suitability of different experimental designs for use with ML models for data analysis and prediction. Consequently, the most promising predictive models will also be indicated. This research is subject to some constraints, that come mainly from the literature background as presented in section \ref{S:2}:
\begin{itemize}
\item The focus is on the designs, although several ML algorithms will be tested. Some indications will also therefore be provided for the choice of ML model, but solely predictive performance will be considered while crucial aspects such as uncertainty quantification or model interpretability will not be examined.
\item Only the non-sequential DOE case is investigated, despite the sequential design approach being very promising. This is mainly for two reasons: (i) the literature \cite{arbo_peg21SLR} shows that the non-sequential approach is still predominant; (ii) in the sequential experimentation setting there is also the need for one initial dataset to fit the algorithms in the first iteration. Since undesigned data are often presently employed as the initial dataset \cite{arbo_peg21SLR}, sequential experimentation will also benefit from the results of this study.
\item The focus is on physical experiments. Consequently:
\begin{itemize}
\item Non-linear test functions that emulate physical processes will mainly be selected.
\item Small data will be considered.
\item Designs with a moderate number of factor levels will be favored.
\item Both the homoscedastic and heteroscedastic noise cases will be taken into consideration.
\end{itemize}
\end{itemize}

\begin{figure}[htbp]
\centering\includegraphics[width=.95\linewidth]{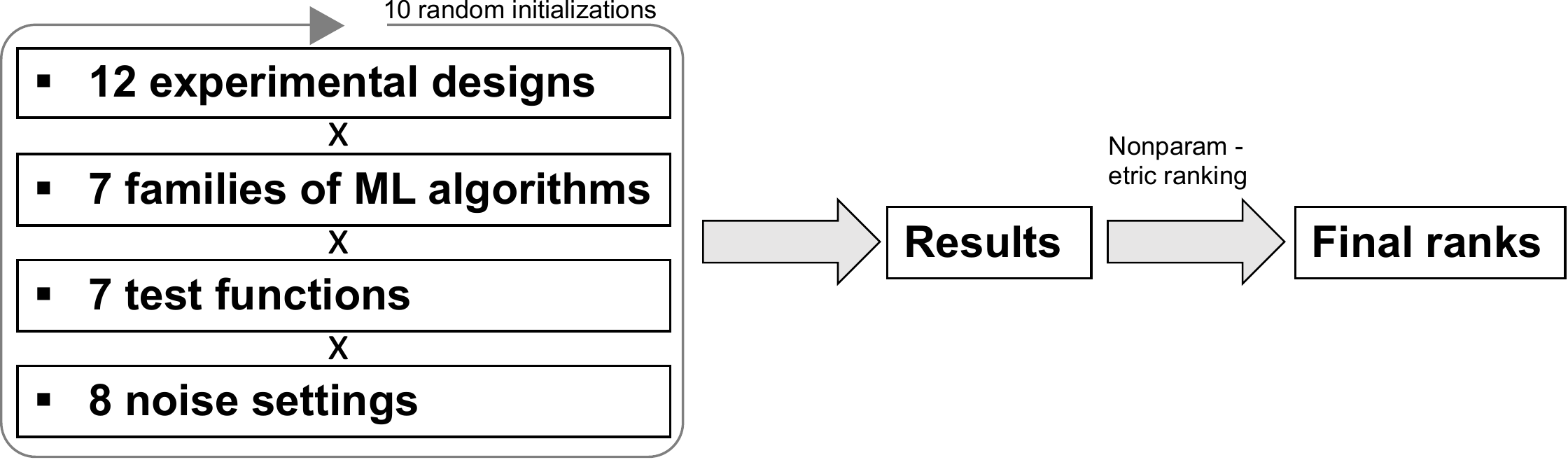}
\caption{General framework of the simulation study.}
\label{sim_framework}
\end{figure}

To the best of our knowledge, this is the first study devoted to the analysis of the impact of choice of experimental design when the method chosen for data analysis is of the ML type. Similar studies have been published that focus on the methods employed for data analysis, but they refer to the case of computer experiments, thus select only space-filling designs for data collection and choose GP models for prediction \cite{chen2016analysis}. In this study we take a larger number of designs with heterogeneous characteristics, and also test a large number of different models. A representation of the general structure of the simulation study is reported in Figure \ref{sim_framework}, and further details are provided in the following sections.

\subsection{Experimental designs}
\label{S:3.2}
Table \ref{DOEs_SS} shows a summary of the experimental designs selected for this study. The designs will be used to collect data on 7 different test functions, all having 6 active dimensions (section \ref{S:3.4}). This is in line with most industrial applications of DOE, as it has been shown that 75\% of case studies in engineering consider 6 or less factors \cite{ilzarbe2008practical}.

Since the principal aim of this work is to provide guidelines to practitioners regarding what experimental designs to choose in physical experimentation when data is analyzed with predictive models, we decided to allocate for each alternative design the same budget in terms of number of trials, to provide a fair comparison between the competitive options. Often this is also the case for practical applications of DOE in industry, where a given amount of resources is allocated for a specific study and the issue is to choose one experimental design to appropriately pursue the goals of the project. For this reason, some powerful designs such as definitive screening designs \cite{jones2011class} are not considered in this study, as they would require a different number of trials given the chosen number of experimental factors, making the comparison with other designs unfair.

Furthermore, and as already pointed out, we suppose that the main objective of the project is to obtain accurate predictions about a specific physical phenomenon. Even though we recognize that other aspects (e.g. uncertainty quantification, interpretability) are also important, we argue that if the analyst selects an intricate predictive model (e.g. a ML algorithm) to analyze a non-linear phenomenon, the main objective is to produce accurate predictions. Some evidence from the literature \cite{arbo_peg21SLR} also confirms this, given that in product innovation ML models are employed for the analysis of DOE data mainly for their advantages in terms of predictive performance and ability to model complex non-linear relationships.

The designs considered in this work have been grouped into three categories: (i) \enquote{classical designs}, (ii) \enquote{optimal designs} and (iii) \enquote{space-filling designs}. In the \enquote{classical designs} category we include the CCD, BBD and FFD. Even though such designs have been around for decades, it has been shown that they are still the most used for data collection in ML studies on product innovation \cite{arbo_peg21SLR}. The \enquote{optimal designs} included are the D-optimal and I-optimal that have recently replaced classical designs in many applications. The designs from the \enquote{space-filling} category have been included to provide a benchmark for those practitioners that may be familiar with the literature on computer experiments, and may therefore want to assess the potential of the predictive models to provide accurate predictions when a maximum number of levels is included in the design. This is because a large number of levels provides more supporting points to the predictive models, and may thus favour a better understanding of the non-linearities in the data.  However, we recognize that such designs are not practical in physical experimentation, and that they are usually employed to analyze the behavior of deterministic computer solvers, while here the main focus is on noisy physical experiments. This is also why only a few designs in this category are selected, and other widely used designs (e.g. maximin LHDs \cite{morris1995exploratory}), are not considered here. Nevertheless, one \enquote{hybrid} design from the space-filling literature with characteristics that enable its application to physical experiments is also considered.

In the following sections we provide details of the designs in each category.

\subsubsection{Classical designs}
Three classical designs have been selected from the RSM literature, namely CCD, BBD and FFD. The CCD was built from a two-level factorial design of resolution V with 4 center points added; an appropriate alpha parameter was then chosen to obtain a rotatable design and axial (star) points were included accordingly. The whole design was scaled to have the range determined by the axial points equal to the range of the levels for each test function, as detailed in section \ref{S:3.4}. In total, 52 runs were required to generate the entire design, and the same budget in terms of runs was selected for all the other designs. Similarly, the BBD includes 4 central points and has 52 runs. Both the CCD and BBD have 3 levels for each factor \cite{CCD,BBD}. The FFD  was built as a full factorial design including 6 levels for each factor. To fit the budget of 52 runs, the design was then computer-optimized using the D-optimality criterion. The rationale behind the choice of number of factor levels for the FFD was that we expected a larger number of levels to provide more granularity to the predictive models, thus enabling a more precise estimate of the underlying non-linear function. Furthermore, we argue that this number of levels is not prohibitive for application in physical experiments, and can represent an appropriate trade-off between a better (expected) predictive performance and a marginally more complicated execution of the experiment.

\subsubsection{Optimal designs}
In this paper we consider designs that satisfy the D- and I-optimality criteria. The D-optimality criterion is the most commonly used, and seeks designs that maximize the determinant of the Fisher information matrix, thus ensuring a minimization of the volume of the confidence ellipsoid around the model parameters. The main objective is to obtain a precise estimate of the effects, therefore D-optimal designs can be considered estimation-oriented. If the goal of the study is, as in this case, prediction-oriented, it may be preferable to opt for I-optimal designs, since they minimize the average prediction variance over the experimental region. Nowadays, this is the default criterion in some commercial software \cite{JMP} when the main goal of the DOE study is to obtain a reliable prediction of the response (over all the design space or in the vicinity of the optimal configuration), such as in RSM applications. Furthermore, the evaluation of the prediction variance over the design space is a popular quality criterion for the choice of a DOE \cite{FDS_plot}.

Another well-known prediction-oriented criterion is G-optimality that ensures a minimization of the maximum variance of prediction. However, it has been shown that a minimization of the maximum variance of prediction implies a greater prediction variance over a large part of the region of interest \cite{rodriguez2010generating}. In this study the objective is to minimize the prediction error over the entire design space, thus we discard the G-optimality criterion and prefer the I-optimal case \cite{goos2016optimal}. Other estimation-oriented optimality criteria also exist (e.g.  A-optimality), but given the different objective of this work, we only focus on the D-optimal case as it is the most prominent in the category. For a more detailed discussion of the main optimality criteria, we refer readers to the paper by \citet{jones2020optimal}, pages 370-372.

In the simulation study the optimal designs (D\_opt, I\_opt) were built levering the capabilities of JMP\textsuperscript{\tiny\textregistered}'s Custom Design platform - this employs the coordinate exchange algorithm \cite{coordexchalgor95} over multiple initializations to generate the preferable design given an optimality criterion. The analyst is required to input the levels for each factor (6 equally-spaced levels), the number of runs (52) and a list of effects that it is necessary or desirable to estimate \cite{JMP}.

Finally, the impact of replication is also assessed for the optimal designs. To this end, D- and I-optimal designs were constructed with JMP\textsuperscript{\tiny\textregistered}'s Custom Design platform having a 50\% level of replication (26 base runs replicated twice, D\_opt\_50\%repl and I\_opt\_50\%repl respectively). A full replication of the base design is often performed in DOE studies and it is one of the funding principles of DOE for physical experimentation \cite{montgomery2017design}. Furthermore, on several occasions it has also been shown to provide advantages for stochastic computer experiments when predictive models (especially GPs) are used \cite{anken2010,BOUKOUVALAS20141088,jalali2017comparison, binois2019replication}.

\subsubsection{Space-filling designs}
Two designs have also been selected from the space-filling family, namely a Random Latin Hypercube Design (LHD\_rand) \cite{McKay79, stein1987large} and a Maximum Projection Design (MAXPRO) \cite{joseph2015maximum}. The LHD\_rand design is one of the first and most prominent designs of the space-filling family \cite{McKay79} and is constructed to spread out the experimental points in a way that ensures each of the factors has all proportions of its distribution represented by input values \cite{McKay79}, with the aim of collecting as much information as possible on the design space. The MAXPRO design is a more recent evolution of the LHD and has been shown to have several advantages over the other space-filling designs, the main one being optimization of the projection properties for all subspaces of the factors \cite{joseph2015maximum, Joseph2016}. This is a fundamental concept for both computer and physical experiments when only some of the factors are active, which is often the case due to the sparsity-of-effects principle \cite{li2006regularities,montgomery2017design}. A consequence of the maximum projection property \cite{joseph2015maximum} is that if some of the factors prove to be inactive during the experiment, they may be dropped with no consequences to the space-filling structure of the design, which is also advisable in a physical experiment. However, both the LHD\_rand and MAXPRO designs have a characteristic that limits their applicability to physical experiments: they have as many factor levels as there are runs, since they tend to maximize the spread of the points in the design space. In practical applications in physical experimentation, a large number of levels for each factor may be neither feasible nor advisable, since the presence of noise may obscure some of the active effects. A recent generalization of the Maximum Projection criterion overcomes this problem by enabling MAXPRO designs to deal with multiple types of factors.

In their recent article, \citet{joseph2020designing} updated the MAXPRO designs allowing both categorical and discrete numeric factors to be treated. In their work the authors refer to the case of computer experiments in which some of the variables of a computer simulator may be categorical (e.g. material type) or discrete numeric (the continuous factor can only take some pre-specified levels, e.g. number of flutes in a solid end milling process) \cite{joseph2020designing}. In this article we mainly leverage one of the advantages of the updated MAXPRO designs: if a limited number of levels is identified, the continuous factors can be restricted to the discrete numeric case, meaning the MAXPRO designs are also applicable in physical experimentation \cite{robust2021}. In our view, and for the scope of the present paper, this makes MAXPRO designs with discrete numeric factors (MAXPRO\_dis) hybrid designs: while they are constructed to be essentially space-filling (although only on a limited number of levels), the limited number of levels makes them competitive with RSM designs for physical experimentation \cite{robust2021}. To our knowledge, this is the only class of designs of the space-filling type that can actually be applied to physical experimentation \cite{robust2021} by taking a limited number of levels for each continuous factor. The MAXPRO\_dis design selected for this study counts 6 equally-spaced levels for each factor in the test functions. As such this design is directly comparable to the FFD and optimal designs, having the same budget in terms of levels for the factors.

Two replicated MAXPRO\_dis designs have also been included in the simulation study: one with a 50\% level of replication (26 base runs replicated twice, MAXPRO\_dis\_50\%repl) and the other with a 25\% level of replication (MAXPRO\_dis\_25\%repl), in which the configurations to replicate have been selected randomly. In this case, given the flexibility of the MAXPRO\_dis designs, the authors wanted to evaluate whether a lower level of replication could be advantageous, even though by doing so a proper separation of noise from signal \cite{binois2019replication} could be attempted only at some randomly-selected locations and not at all configurations (as is the case for a 50\% level of replication).

\begin{table}[htbp]
\centering
\caption{Summary of the experimental designs chosen for the simulation study.}
\scalebox{0.6}{
\begin{tabular}{l l l l l l l}
\hline
\textbf{ID} & \textbf{Description} & \textbf{Software} & \textbf{\# levels} & \textbf{\# factors}& \textbf{\# runs} & \textbf{Replication}\\
\hline

CCD & Central Composite Design & R\cite{doewrapper} & 3 & 6 & 52 & 0\% \\
BBD & Box-Behnken Design & R\cite{doewrapper} & 3 & 6 & 52 & 0\% \\
FFD & D-optimal full factorial design & Minitab\textsuperscript{\tiny\textregistered} & 6 & 6 & 52 & 0\%\\
D\_opt & D-optimal design & JMP\textsuperscript{\tiny\textregistered}  & 6 & 6 & 52 & 0\%\\
I\_opt & I-optimal design & JMP\textsuperscript{\tiny\textregistered}  & 6 & 6 & 52 & 0\%\\
LHD\_rand & Random Latin Hypercube Design & R \cite{dicedesign}  & 52 & 6 & 52 & 0\% \\
MAXPRO & MaxPro space-filling design & R \cite{joseph2015maximum}  & 52 & 6 & 52 & 0\%  \\
MAXPRO\_dis & MaxPro discrete numeric design & R \cite{joseph2020designing}  & 6 & 6 & 52 & 0\% \\
D\_opt\_50\%repl & D-optimal design & JMP\textsuperscript{\tiny\textregistered}  & 6 & 6 & 52 & 50\% \\
I\_opt\_50\%repl & I-optimal design & JMP\textsuperscript{\tiny\textregistered}  & 6 & 6 & 52 & 50\% \\
MAXPRO\_dis\_50\%repl & MaxPro discrete numeric design & R \cite{joseph2020designing}  & 6 & 6 & 52 & 50\% \\
MAXPRO\_dis\_25\%repl & MaxPro discrete numeric design & R \cite{joseph2020designing}  & 6 & 6 & 52 & 25\% \\

\hline
\label{DOEs_SS}
\end{tabular}}
\end{table}

\subsection{Machine learning models}
\label{S:3.3}
\citet{arbo_peg21SLR} showed that ANNs are by far the most adopted algorithm for the analysis of data from experimental designs when the purpose is to predict one or more responses. Other models adopted in the literature \cite{arbo_peg21SLR} are Support Vector Regression models (SVMs)\cite{hastie2009elements}, Gaussian Processes (GPs) \cite{gramacy2020surrogates} and linear models (LMs) based on quadratic regression with interactions \cite{montgomery2017design}. In addition to those methods, in this study we investigate the performance of Random Forests (RFs)\cite{breiman2001random} and the algorithms contained within the Automated Machine Learning (aml) platform offered by H2O \cite{ledell2020h2o}, that provides a comprehensive view of the most used ML models.

Careful tuning was carried out to choose the best configuration of the hyperparameters that minimizes the root mean squared error (\texttt{RMSE}) of each algorithm, as detailed below:
 \begin{itemize}
     \item ANN shallow (ANN\_sh): this is the simplest implementation of ANNs, counting one hidden layer. The number of neurons was searched for in the range $[3-12]$, and to limit overfitting, weight decay was applied with a level of $[0-0.5]$. After some initial tuning, linear activation function was selected for the output node. All hyperparameters were optimized by means of fivefold Cross Validation (5-fold CV) \cite{james2013introduction} applied on the training data from the experimental designs as detailed in Table \ref{DOEs_SS}.
     \item ANN deep (ANN\_dp): in this case, two to four hidden layers were considered for the ANN, with either 6 or 12 neurons per layer. Such configurations of the multilayered ANN were identified as the most promising in a preliminary tuning of the models, though a concrete risk of overfitting was identified, especially for the deepest networks. Several activation functions were also tested, including \texttt{Tanh}, \texttt{Maxout} and \texttt{Rectified  Linear} \cite{candel2021deep}. In order to limit overfitting of the training data, dropout \cite{srivastava2014dropout} and both $\ell_1$ (Lasso) and $\ell_2$ (Ridge) regularization were employed \cite{candel2021deep}, making the models more stable and less susceptible to noise. 5-fold CV was used for the selection of the best hyperparameter configuration, as it is the most balanced solution both in terms of computational effort and reduction in variance of the error estimation \cite{james2013introduction}.
     \item Random Forest (RF): 2000 trees were grown from the training data, constituting the RF. This configuration was obtained after an accurate initial tuning, and it has already been used in the literature for application on DOE data \cite{peg_JAS}. The number of candidate predictors selected for each split is chosen through 5-fold CV, as such obtaining decorrelation of the base learners and improved predictions \cite{james2013introduction}.
     \item Support Vector Regression model (SVM): three different kernels were considered for SVM, namely linear $k(\bold{x}, \bold{x'}) =< \bold{x}, \bold{x}'>$, polynomial $k(\bold{x}, \bold{x}') = (\texttt{scale} < \bold{x}, \bold{x}' > +\texttt{offset})$\textsuperscript{\texttt{degree}}, and Gaussian radial basis function (RBF) $k(\bold{x}, \bold{x}') = \texttt{exp} ( - \sigma || \bold{x} -  \bold{x}' ||^2)$ where $\bold{x}$ and $\bold{x}'$ are two input vectors \cite{kernlab}. 5-fold CV was employed for tuning the hyperparameters (e.g. $\texttt{scale}$, $\texttt{offset}$ and $\texttt{degree}$ in the polynomial kernel or $\sigma$ in the RBF kernel).
    \item Gaussian Process (GP): it is usually the model of choice for the analysis of data from computer experiments, typically fitting data from space-filling designs \cite{gramacy2020surrogates}. Consider a $d$-dimensional design $\bold{D}$ with $n$ runs $\bold{D}=\{\bold{x}_1, ...,\bold{x}_n\}$, where $\bold{x}=(x_1,...,x_d)$ denotes one of the input vectors and $\bold{y}=(y(\bold{x}_1), ...,y(\bold{x}_n))^{T}$ is the vector of the outputs. The GP assumption states that $y(\bold{x})$ is a realization of:
    \begin{equation}
        Y(\bold{x})=\mu(\bold{x})+Z(\bold{x})
    \end{equation} 
    where $\mu(\bold{x})$ is the trend component and $Z(\bold{x})$ is multivariate normal with mean 0 and covariance function $k$ (kernel function): 
    \begin{equation}
        Z(\bold{x}) \sim \mathcal{N}(0,\,\tau^2 k(\cdot,\cdot))
    \end{equation}
    where $\tau^2$ is a multiplicative constant known as the scale parameter, defining the amplitude of the function \cite{gramacy2020surrogates}. The choice of kernel function $k(\cdot,\cdot)$ is critical, and many different options have been proposed in the literature. For an extensive discussion on the properties of the commonly used covariance functions for GPs, please refer to the work by \citet{GP4ML2006}, chapter 4.

    Let ${x}$ and ${x}'$ denote two components of the input vectors  $\bold{x}$ and $\bold{x}'$ respectively; in this article we consider the following kernel functions \cite{roustant2012dicekriging}: 
    
    \begin{itemize}
        \item Gaussian: $k({x},{x}')=\texttt{exp}(-\frac{({x}-{x}')^2}{2 \theta^2})$
        \item Exponential: $k({x},{x}')=\texttt{exp}(-\frac{|{x}-{x}'|}{\theta})$
        \item Power-Exponential: $k({x},{x}')=\texttt{exp}(-(\frac{|{x}-{x}'|}{\theta})^t)$, with $0<t\leq2$
        \item Matérn $5/2$: $k({x},{x}')=(1+\frac{\sqrt{5} |{x}-{x}'|}{\theta}+ \frac{5 ({x}-{x}')^2}{3 \theta^2})\texttt{exp}(-\frac{\sqrt{5} |{x}-{x}'|}{\theta})$
        \item Matérn $3/2$: $k({x},{x}')=(1+\frac{\sqrt{3} |{x}-{x}'|}{\theta})\texttt{exp}(-\frac{\sqrt{3} |{x}-{x}'|}{\theta})$
    \end{itemize}

 Note that here the kernel functions are reported on the 1-$d$ case, but they can be generalized to the $d$-dimensional case by taking $K(\bold{x},\bold{x}')= \prod\limits_{i=1}^{d} k({x}_i,{x}_i')$ \cite{roustant2012dicekriging}. $\theta$ is the lengthscale parameter which defines the rate of decay of the spatial correlation among two data points; in this article, we consider the vectorized version of $\bm{\theta}=(\theta_1,...,\theta_d)$, that allows the strength of correlation to be modulated separately by distance in each input coordinate \cite{gramacy2020surrogates}. Both $\bm{\theta}$ and $\tau^2$ are estimated via maximum likelihood estimation (MLE) \cite{roustant2012dicekriging}. Another relevant parameter is the \enquote{nugget} $g$, whose main practical role is to prevent problems of inversion of the covariance matrix due to the presence of numerical instabilities \cite{roustant2012dicekriging,gramacy2020surrogates}, but can also provide protection from violations of the modeling assumptions (e.g. inappropriate covariance function) or other practical occurrences such as data sparsity attributable to high dimensionality and small data sizes \cite{gramacy2012cases}. The nugget effect is added to the diagonal terms of the covariance matrix, and in this article $g=10\textsuperscript{-8} \texttt{var}(\bold{y})$ is chosen.

    The choices for the trend component $\mu(\bold{x})$ are either:
    \begin{itemize}
        \item Constant, with only the intercept $\beta_0$.
        \item Quadratic with interactions and stepwise selection of significant terms.
    \end{itemize}

    \item Automated Machine Learning (aml): this is a user-friendly ML software offered by H2O. Its main advantage is that it automates the ML workflow, which includes automatic training and tuning of many models by 5-fold CV and within a user-specified time limit \cite{amlweb}. The default algorithms evaluated by aml include: RFs, Extremely Randomized Trees, Generalized Linear Models, Gradient Boosting Machines, Deep Neural Networks and Stacked Ensembles of all the base learners \cite{cook2016practical, landry2020machine}.
    
    \item Second-order model \cite{montgomery2017design} and stepwise selection of relevant effects (LM):
    \begin{equation}
        y=\beta_{0}+\sum_{i=1}^{d}{\beta_{i}x_{i}}+\sum_{i=1}^{d}{\beta_{ii}x_{i}^{2}}+\mathop{\sum\sum}_{i<j}{\beta_{ij}x_{i}x_{j}} + \epsilon
    \label{LM}
    \end{equation}
    where $\beta_{0}$, $\beta_{i}$, $\beta_{ii}$ and $\beta_{ij}$ are the regression coefficients and $x_{i}$, $x_{j}$ are the input variables with $i, j = 1,\dots,d$.

 \end{itemize}

\subsection{Test functions}
\label{S:3.4}
The test functions were selected from the open literature \cite{simulationlib}, mainly emulating physical processes. All the functions were restricted to 6 active dimensions and are evaluated on the training data, i.e. the experimental designs (Tab. \ref{DOEs_SS}), and on a noiseless random LHD with 1000 points that is used as test data \cite{chen2016analysis}. In the study the dependent variables were standardized: 

\begin{equation}
y_n\textsuperscript{std}=\frac{y_n-\overline{y}}{\sigma_y}
\label{zscore}
\end{equation}

where $y_n\textsuperscript{std}$ is the standardized value and $y_n$ is the observed value for the $n$-th observation, $\overline{y}$ and $\sigma_y$ are the mean and standard deviation of $y$ respectively. For each test function, $\overline{y}$ and $\sigma_y$ were estimated from 100 random LHDs each with $500000$ data points: $\overline{y}=(\sum_{i=1}^{100}\overline{y}_{LHD_i})/100$ where $\overline{y}_{LHD_i}$ is the mean response over the $i$-th design, and  $\sigma_y=\max \sigma_{y LHD_i}$ with $\sigma_{y LHD_i}$  the standard deviation of the response calculated over the $i$-th design ($i=1,...,100$). Independent variables are normalized to $0-1$. A summary of the test functions is reported in Table \ref{test_fnct}, and further details are provided in supplemental material.

\begin{table}[htbp]
\centering
\caption{Summary of the test functions chosen for the simulation study.}
\scalebox{0.65}{
\begin{tabular}{l l l l l l}
\hline
\textbf{ID} & \textbf{Description} &  \textbf{\# dimensions}\\
\hline

Borehole & models water flow through a borehole & 6 \\
OTL circuit & models an output transformerless push-pull circuit & 6 \\
Piston & models the circular motion of a piston within a cylinder & 6 \\
Piston Mod & a modification of Piston function, with increased non-linearity & 6 \\
Robot arm & models the position of a robot arm which has 3 segments & 6 \\
Rosenbrock Function & is a popular test problem for optimization algorithms & 6\\
Wing weight & models a light aircraft wing & 6 \\

\hline
\label{test_fnct}
\end{tabular}}
\end{table}

\subsection{Noise levels}
\label{S:3.5}
\begin{figure}[htbp]
\centering\includegraphics[width=1\linewidth]{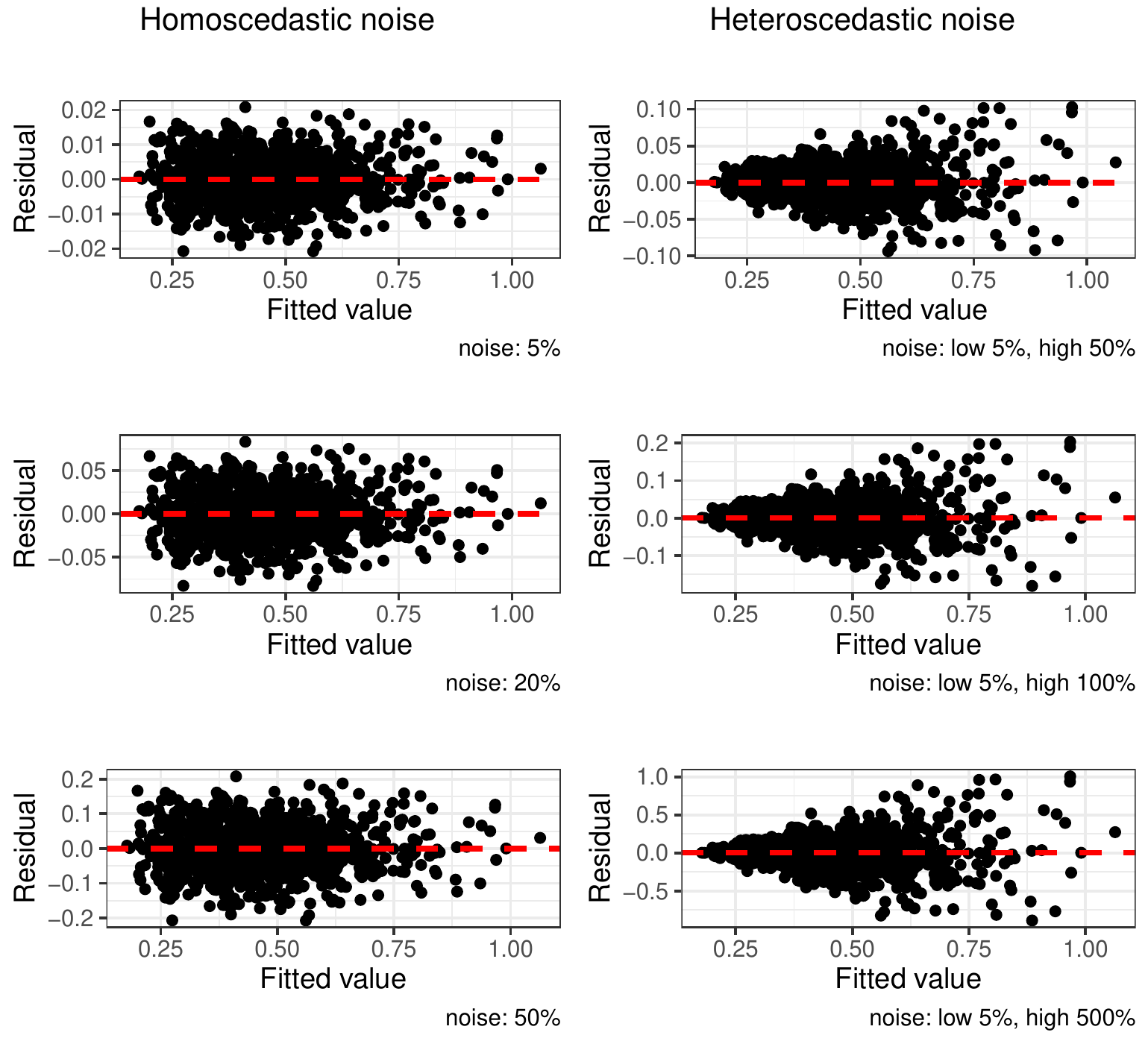}
\caption{Visualization of the impact of different noise structures and magnitudes, both homoscedastic (on the left) and heteroscedastic (on the right) for the test function \enquote{Piston}. The residuals are computed as the difference between the deterministic function and the noisy function on a 1000-point LHD. Similar results are obtained for the other test functions.}
\label{noisefig}
\end{figure}

In this study we investigate both the homoscedastic and heteroscedastic noise cases. For the homoscedastic case, we assume a random normal noise component $\epsilon \sim \mathcal{N}(0,\sigma_{hom}^2)$, with $\sigma_{hom}$ in the range $[0, 0.5\sigma_y]$ \cite{picheny2013benchmark}. For the heteroscedastic case, we assume a random normal error $\epsilon \sim \mathcal{N}(0,\sigma_{het}^2)$ that increases linearly with the response. That is, at a specific design configuration $\bold{x}$: $\sigma_{het}=a(f(\bold{x})+b)$, with $a=(\texttt{m}\sigma_y-0.05\sigma_y)/(\max y-\min y)$, $b=(0.05\sigma_y-a\min y)/a$, where $\min y$ and $\max y$ are the minimum and maximum $y$ registered over the 100 LHDs respectively (section \ref{S:3.4}) and $\texttt{m}$ is in $[0.5, 5]$ depending on the magnitude of noise \cite{jalali2017comparison}. A summary of the different noise levels is reported in Table \ref{noisetab}, and the impact of different noise magnitudes is visualized in Figure \ref{noisefig}.

\begin{table}[htbp]
\centering
\caption{Summary of the noise structures chosen for the simulation study.}
\scalebox{0.65}{
\begin{tabular}{l l l l }
\hline
\textbf{Noise} \boldsymbol{$\sigma$} & \textbf{Type} &  \textbf{Description}\\
\hline

$\sigma_{hom}=0\sigma_{y}$ & -- & 0\% noise, deterministic function \\
$\sigma_{hom}=0.05\sigma_{y}$ & Homoscedastic & 5\% noise \\
$\sigma_{hom}=0.125\sigma_{y}$ & Homoscedastic & 12.5\% noise \\
$\sigma_{hom}=0.2\sigma_{y}$ & Homoscedastic & 20\% noise\\
$\sigma_{hom}=0.5\sigma_{y}$ & Homoscedastic & 50\% noise \\
$\sigma_{het,min}=0.05\sigma_{y}$,  & \multirow{ 2}{*}{Heteroscedastic}   & Moderate: 5\% noise at $\min y$\\
$\sigma_{het,max}=0.5\sigma_{y}$ & & and 50\% noise at $\max y$   (low5\_high50)\\
$\sigma_{het,min}=0.05\sigma_{y}$,  & \multirow{ 2}{*}{Heteroscedastic}   & Intermediate: 5\% noise at $\min y$  \\
$\sigma_{het,max}=1\sigma_{y}$ & & and 100\% noise at $\max y$ (low5\_high100)\\
$\sigma_{het,min}=0.05\sigma_{y}$,  & \multirow{ 2}{*}{Heteroscedastic}   & Severe: 5\% noise at $\min y$  \\
$\sigma_{het,max}=5\sigma_{y}$ & & and 500\% noise at $\max y$ (low5\_high500)\\

\hline
\label{noisetab}
\end{tabular}}
\end{table}

\subsection{Ranking procedure}
\label{S:3.6}
Given the large number of designs, models, test functions and noise structures and magnitudes, it should already be clear that a definitive ranking of the DOEs and models with respect to the prediction error obtained on a test dataset would be quite difficult to achieve. As such, a robust ranking procedure that can give strong and reliable insights on the results is required, and for this article we select a procedure based on the execution of nonparametric permutation tests. The main advantages of this procedure include high flexibility and the possibility of applying it without requiring any strict assumption on the size or underlying distribution of data, while nonetheless providing inferential information \cite{pesarin2010permutation}. Other ranking approaches have been proposed in the literature \cite{gupta2002}, mainly parametric. However, in comparison to the approach considered here those methods are more demanding from the point of view of statistical assumptions (e.g. about data distribution) that cannot always be verified in practical applications. Furthermore, other ranking and selection procedures need to specify the probability of correct classification and how far two ordered populations should be from each other in the ranking \cite{gupta2002}, requirements that are not needed in the approach adopted here \cite{arboretti2014permutation,rankbook16}. Additionally, the considered nonparametric ranking procedure also proved its efficacy from the point of view of the reliability of results as it has been extensively validated by means of simulation studies \cite{arboretti2014permutation} and practical applications considering both observational and experimental data in several fields \cite{rankbook16}, including medicine, new product development and marketing studies. Recently it has also been included as a core component in a methodology that performs variable selection in near-infrared spectroscopy using ML models\cite{arboretti2021interval}.

In this paper the permutation tests are applied using the difference in means as test statistics and assuming independent or paired data, depending on the specific situation. Considering $G_i$ and $G_j$ with $i,j=1,...,C, i \neq j$ two different groups of data to be compared (e.g. the different experimental designs), the permutation testing framework is employed to test the directional alternative hypothesis $\texttt{RMSE}_{G_i}>\texttt{RMSE}_{G_j}$, where $\texttt{RMSE}$ is the prediction error calculated on the test data, i.e. a noiseless random LHD with 1000 data points \cite{chen2016analysis}. 2000 permutations are considered and the relevant p-values are computed by means of the Conditional Monte Carlo (CMC) procedure described in \citet{pesarin2010permutation}. In this study we refer to the one-way MANOVA layout (although in a nonparametric framework), meaning that interest lies in the detection of differences between $C$ groups on $p$ dependent variables (e.g. the test error obtained with different predictive models). Therefore, a multivariate ranking procedure is needed and we opt for the permutation approach for ranking multivariate populations described by \citet{arboretti2014permutation}. After the execution of all possible comparisons between the $C$ groups over the $p$ dependent variables and the application of Fisher's combining function as suggested in \cite{pesarin2010permutation, arboretti2014permutation} for the combination of the p-values over the $p$ variables, matrix $\bold{P}$ with dimension $C\times C$, that includes the combined p-values related to the comparisons between the $C$ groups, is obtained.

The ranking procedure  \cite{arboretti2014permutation} is as follows:
\begin{enumerate}
    \item Generate the matrix $\bold{S}$ where $S_{ij}=1$ if ${P}_{ij} \leq \alpha /2$, $S_{ij}=0$ otherwise. ${P}_{ij}$ is an entry of $\bold{P}$.
    \item Compute the vector $\bold{r}^D$ of downward rank estimates $r_j^D=1+\sum_{i=1}^C S_{ij}$, $j=1,...,C$.
    \item Compute the vector $\bold{r}^U$ of upward rank estimates $r_i^U=1+\{\# (C-\sum_{j=1}^C S_{ij}) > (C-\sum_{j=1}^C S_{i'j}), i'=1,...,C, i' \neq i\}$, $i=1,...,C$.
    \item Compute the vector $\bold{r}$ of ranking estimates $r_i=1+\{\#(r_i^D+r_i^U)/2>(r_j^D+r_j^U)/2, j=1,...,C, i \neq j\}, i=1,...,C$.
\end{enumerate}

In the next section we will present the final rankings for both the DOEs and ML models considered in this work. In order to achieve a final ranking of the different groups, two applications of the ranking procedure described above will be required.

Consider the case in which the interest lies in ranking the experimental designs. In the first application of the ranking procedure, both the test function and noise setting will be fixed, meaning that a ranking of the designs for each specific test function in a particular noise setting will be obtained. In this case the $C$ groups refer to the experimental designs (Tab. \ref{DOEs_SS}), while the $p$ dependent variables are the $p$ vectors of test errors obtained with the $p=7$ predictive models (section \ref{S:3.3}). The groups $G$ are assumed to be independent. For each of the $p$ variables, each group $G$ will have 10 observations, that are 10 values of \texttt{RMSE} on the test data obtained with 10 repetitions of the simulation in which the 5-fold CV procedure and the predictive models are initialized with a different random seed.

At this point, a second application of the ranking procedure is performed, taking advantage of the results obtained in the first iteration. In this case, the noise levels will remain fixed, and the ranking of the designs will be performed over the different test functions, meaning that the $C$ groups will still refer to the designs, but now each $G$ will have 7 observations that are the ranks of the designs from the first iteration obtained on the 7 test functions given a specific noise setting. In this case $p=1$, meaning that the univariate version of the method in \citet{arboretti2014permutation} is used. The final result consists in an overall rank of each group $G$ for each specific noise setting.

When interest lies in achieving a rank of the predictive models, the procedure is similar: the $C$ groups will refer to the 7 predictive models (section \ref{S:3.3}), and the $p$ dependent variables are the $p$ vectors of test errors obtained by fitting the algorithms on data from the $p=12$ experimental designs (Tab. \ref{DOEs_SS}). In this case the groups $G$ cannot be assumed independent since for a given experimental design all the models are fitted to the same data. In all hypothesis testing procedures, $\alpha$ is assumed to be $5\%$.

\section{Results}
\label{S:4}
This section presents the results of the simulation study and provides the intermediate and final ranks for both the experimental designs and predictive algorithms; sections \ref{S:4.1} and \ref{S:4.2} respectively.

\subsection{Results for the experimental designs}
\label{S:4.1}
\begin{figure}[htbp]
\centering\includegraphics[width=1\linewidth]{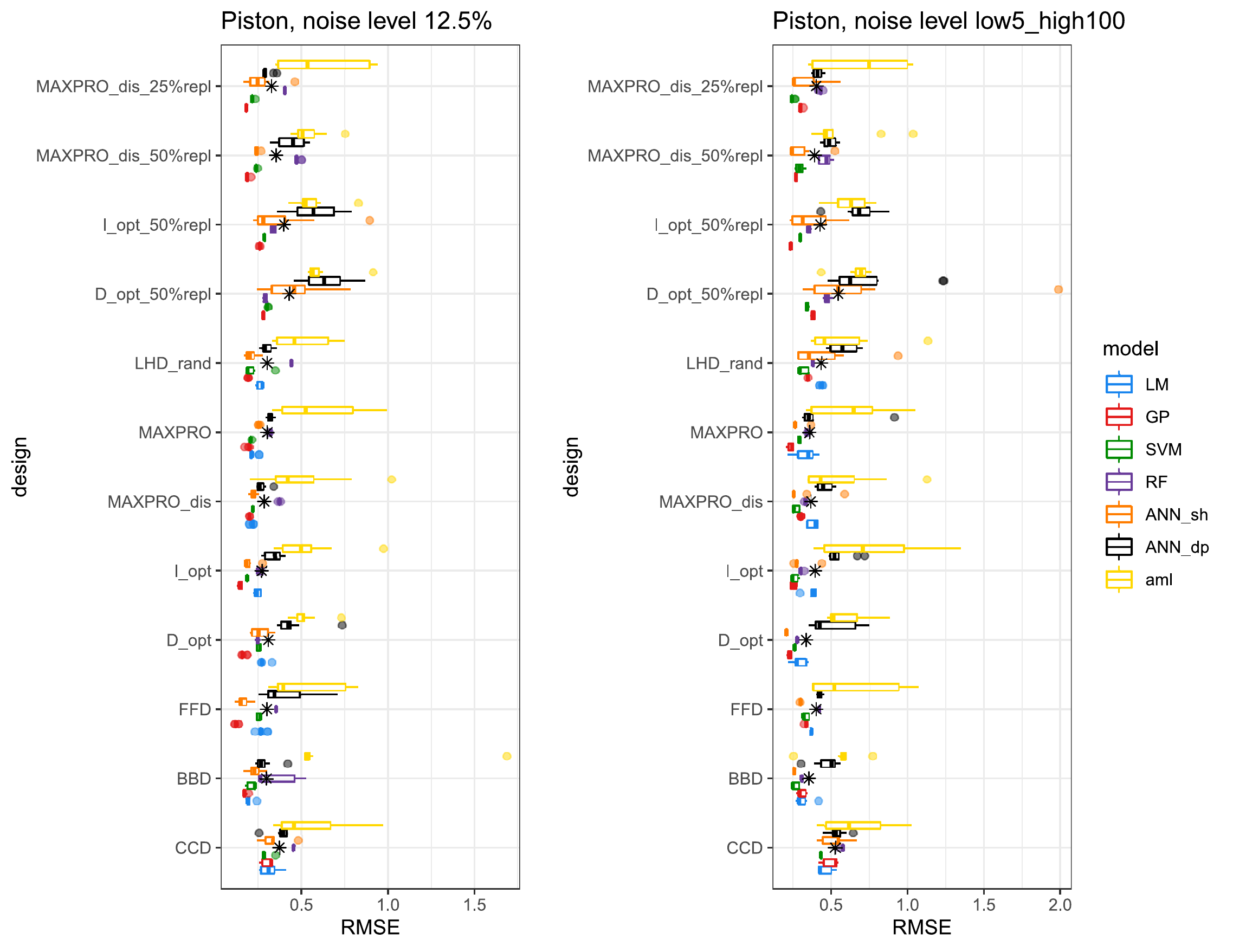}
\caption{Results of the 10 iterations of the simulation study for the function \enquote{Piston}, considering the test error (\texttt{RMSE}) obtained by the different models trained on the different experimental designs for an intermediate level of homoscedastic (left) and heteroscedastic (right) noise. The mean value for each group is also indicated by a black star. Similar results are obtained for the remaining test functions and noise settings.}
\label{bplot_piston}
\end{figure}

Figure \ref{bplot_piston} shows the results in terms of the test error obtained by the models trained on the data collected through the different experimental designs. For the sake of brevity, the figure focuses on the results of the test function \enquote{Piston}, considering only one level of error for both the homoscedastic and heteroscedastic cases. Similar results are obtained for the other noise levels and test functions, while synthetic information will be reported in the following sections for all test functions and noise levels. It should be noted that for replicated designs, it was not possible to fit the LM model, since too few unique data points were present in the training data to allow the estimation of parameters.

\subsubsection{First ranking}
\label{S:4.1.1}
At this stage the first application of the multivariate ranking procedure described in section \ref{S:3.6} is conducted. Figure \ref{1st_rank_piston} (Appendix A) shows the results for the function \enquote{Piston} considering all the noise settings. The ranking is conducted for each test function, and a summary of the results is displayed in Figures \ref{rankDES_allnoise_hom} and \ref{rankDES_allnoise_het}, with each plot referring to a specific noise setting.

\subsubsection{Final ranking}
\label{S:4.1.2}
The following step consists in the execution of an additional iteration of the ranking procedure as described in section \ref{S:3.6}. The final rank provides a synthetic result for the intermediate rankings displayed in each panel of Figures \ref{rankDES_allnoise_hom} and \ref{rankDES_allnoise_het}, essentially carrying out a ranking of the ranks from the first iteration. Results are displayed in Tables \ref{tab_rank_DOE_hom} and \ref{tab_rank_DOE_het} for the homoscedastic and heteroscedastic cases respectively. Each table should be read column-wise: the relative ranks are computed for each noise setting.

\subsection{Results for the predictive models}
\label{S:4.2}
The procedure for ranking the predictive models is more or less equivalent to the one followed for the experimental designs, with slight modifications as described in section \ref{S:3.6}. Figures \ref{rankMOD_allnoise_hom} and \ref{rankMOD_allnoise_het} (Appendix B) show a summary of the intermediate ranks obtained for each test function and noise setting, both homoscedastic and heteroscedastic. The results of the final ranks are shown in Tables \ref{tab_rank_MODEL_hom} and \ref{tab_rank_MODEL_het}.

\begin{table}
\centering
\caption{Final rank of the designs for the homoscedastic noise cases.}
\scalebox{0.75}{
\begin{tabular}{l c c c c c l }
\cline{1-6}
\textbf{Design} & \textbf{0\%} &  \textbf{5\%} &\textbf{12.5\%} & \textbf{20\%} &  \textbf{50\%} & \\
\cline{1-6}
CCD	&		8	&		9	&		11	&		10	&		12		&\\
BBD	&		6	&		6	&		4	&		4	&		4		&\\
FFD	&		5	&		1	&		1	&		1	&		1		&\\
D\_opt	&		9	&		8	&		8	&		7	&		4	&	\\
I\_opt	&		3	&		1	&		4	&		2	&		1	&	\\
MAXPRO\_dis	&		1	&		1	&		1	&		2	&		4&		\\
MAXPRO	&		2	&		1	&		1	&		4	&		1		&\\
LHD\_rand	&		3	&		1	&		6	&		6	&		8	&	\\
D\_opt\_50\%repl	&		12	&		12	&		12	&		12	&		11	&	\\
I\_opt\_50\%repl	&		11	&		11	&		10	&		10	&		4	&	\\
MAXPRO\_dis\_50\%repl	&		10	&		9	&		9	&		9	&		8 &		\\
MAXPRO\_dis\_25\%repl	&		7	&		7	&		7	&		8	&		8 &		\\

\cline{1-6}

\label{tab_rank_DOE_hom}
\end{tabular}}
\end{table}

\begin{table}
\centering
\caption{Final rank of the designs for the heteroscedastic noise cases. }
\scalebox{0.75}{
\begin{tabular}{l c c c  l}
\cline{1-4}
\textbf{Design} & \textbf{low5\_high50} &  \textbf{low5\_high100} &\textbf{low5\_high500} & \\
\cline{1-4}

CCD	&							11	&		11	&		11	&	\\
BBD	&							3	&		4	&		1		&\\
FFD	&							5	&		7	&		8	&	\\
D\_opt	&						8	&		4	&		8	&	\\
I\_opt	&						3	&		1	&		6	&	\\
MAXPRO\_dis	&					2	&		1	&		1	&	\\
MAXPRO	&						1	&		3	&		6	&\\
LHD\_rand	&					6	&		7	&		8	&	\\
D\_opt\_50\%repl	&			11	&		11	&		12	&	\\
I\_opt\_50\%repl	&			10	&		9	&		1	&	\\
MAXPRO\_dis\_50\%repl	&		9	&		9	&		1	&	\\
MAXPRO\_dis\_25\%repl	&		6	&		4	&		1	&	\\

\cline{1-4}

\label{tab_rank_DOE_het}
\end{tabular}}
\end{table}

\begin{table}
\centering
\caption{Final rank of the models for the homoscedastic noise cases.}
\scalebox{0.75}{
\begin{tabular}{l c c c c c l }
\cline{1-6}
\textbf{Model} & \textbf{0\%} &  \textbf{5\%} &\textbf{12.5\%} & \textbf{20\%} &  \textbf{50\%} & \\
\cline{1-6}

LM	&			2	&		2	&		2	&		2	&		3	&		\\
GP	&			1	&		1	&		1	&		1	&		1	&		\\
SVM	&			4	&		3	&		2	&		3	&		2	&		\\
RF	&			5	&		5	&		5	&		5	&		5	&		\\
ANN\_sh	&		3	&		3	&		4	&		4	&		3	&		\\
ANN\_dp	&		7	&		7	&		5	&		7	&		7	&		\\
aml	&			6	&		5	&		5	&		5	&		6	&		\\

\cline{1-6}

\label{tab_rank_MODEL_hom}
\end{tabular}}
\end{table}

\begin{table}
\centering
\caption{Final rank of the models for the heteroscedastic noise cases. }
\scalebox{0.75}{
\begin{tabular}{l c c c l }
\cline{1-4}
\textbf{Model} & \textbf{low5\_high50} &  \textbf{low5\_high100} &\textbf{low5\_high500} & \\
\cline{1-4}

LM	&			3	&		4	&		4	&		\\
GP	&			1	&		1	&		3	&	\\
SVM	&			1	&		2	&		1	&		\\
RF	&			5	&		5	&		2	&	\\
ANN\_sh	&		4	&		3	&		6	&		\\
ANN\_dp	&		6	&		7	&		6	&		\\
aml	&			6	&		6	&		5	&		\\

\cline{1-4}

\label{tab_rank_MODEL_het}
\end{tabular}}
\end{table}

\section{Discussion}
\label{S:5}
In this section we discuss the results of the simulation study, focusing both on the experimental designs (section \ref{S:5.1}) and predictive models (section \ref{S:5.2}).

\subsection{Discussion: experimental designs}
\label{S:5.1}
In general, and as expected, the results of the study show that the choice of experimental design impacts on the final outcome of the analysis. From the unprocessed results (e.g. Fig. \ref{bplot_piston}) it can already be seen that not all the designs perform equally, and some trends can already be detected (e.g. for intermediate noise magnitudes the replicated designs do not offer an advantage). However, the raw results clearly indicate the need for the application of a robust ranking procedure to provide valuable synthetic information.

The first application of the ranking procedure provides some preliminary indications about the experimental designs. Given a test function (e.g. Fig \ref{1st_rank_piston}), the presence of noise largely affects the efficacy of the designs: in general, for modest levels of noise, the designs with many factor levels appear to be favored, while as  noise increases, the performance of designs with a moderate number of levels and the addition of replicates increases. Comparing the ranks of the designs for the different test functions appears to confirm this trend (Fig. \ref{rankDES_allnoise_hom}, \ref{rankDES_allnoise_het}). Furthermore, an interaction effect seems to exist between the test functions and the experimental designs because, in general, notwithstanding the level of noise, some designs tend to perform better for some test functions than others: take for example the functions \enquote{Robot arm} and \enquote{Rosenbrock} for which the designs with many factor levels tend to consistently perform better than designs with only three factor levels. We put this down to the different degrees of non-linearity of each function. However, as the magnitude of noise increases, the situation progressively becomes less clear and for severe levels of noise, whether homoscedastic or heteroscedastic, it becomes burdensome for the analyst to choose by simply considering the results of the first rank.

The results of the second rank provide clearer indications, and confirm that not all the designs perform equally as well. Focusing on the homoscedastic case (Tab. \ref{tab_rank_DOE_hom}), FFD, MAXPRO\_dis and MAXPRO are the overall best performers, followed closely by the I-optimal design. BBD and LHD\_rand follow, with the BBD being the best of the designs with only three levels for each factor. The D-optimal design and the CCD are among the worst performers. In general, the presence of replicates makes the predictions worse, but a lower level of replication is preferable to a larger one.

The situation differs when focus turns to the heteroscedastic noise setting (Tab. \ref{tab_rank_DOE_het}). In this case, the MAXPRO\_dis is the unquestionable winner, ranking first for the intermediate and severe noise situations, and second in the case of moderate heteroscedasticity. The BBD, I\_opt, MAXPRO and MAXPRO\_dis\_25\%repl follow, all behaving similarly. The MAXPRO\_dis\_50\%repl, D\_opt, FFD, I\_opt\_50\%repl and LHD\_rand perform less well, and the ranking is completed by the CCD and D\_opt\_50\%repl designs.

From the analysis it emerges that the MAXPRO\_dis is the overall best performer if both the homoscedastic and heteroscedastic settings are considered. Apparently, the space-filling property of the design can propose combinations of the factor settings that maximize the ability of several different predictive models to capture the non-linearity inherent in the phenomenon. At the same time, the reduced number of factor levels makes the design robust to the presence of noise and applicable for physical experimentation. The MAXPRO\_dis significantly outperforms all the other designs with 6 levels for the factors, namely FFD, D\_opt and I\_opt. Interestingly, while such designs perform similarly (except for D\_opt) in the homoscedastic setting, in the heteroscedastic scenario the MAXPRO\_dis significantly outperforms its competitors despite having the same number of factor levels. Particularly relevant is the fact that not only the FFD and estimation-oriented D-optimal designs are outperformed, but also the I-optimal design that is specifically built to favor accurate predictions of the unknown underlying function within the design space. The specific reasons for the outstanding performance of the  MAXPRO\_dis are still unknown, but the authors believe that they are related to the intrinsic space-filling criterion adopted by the designs. Confirmation of the merit of such designs is that the traditional space-filling version of the MAXPRO also outperforms its direct competitor, i.e. the LHD\_rand design, both in the homoscedastic and heteroscedastic scenarios. While the LHD\_rand design achieves good results only for rather small noise levels, the MAXPRO design proves to be quite effective in intermediate and large noise situations as well. However, it should again be stressed that traditional space-filling designs are not a viable solution in many practical applications of physical experimentation, and are reported here only as a benchmark. Furthermore, as expected from the optimal designs, the prediction-oriented I\_opt design consistently outperforms the estimation-oriented D\_opt option. This implies that even though the considered optimal designs have historically been developed to be used together with parametric second-order models (LM), their properties also appear to be retained when the model for data analysis is from the ML family.

Another surprise is the difference between the CCD and BBD. These classic designs both have three levels for each factor, but in general the BBD performs better, especially in settings with large noise in which it is among the overall best performers. Apparently, the edge points of BBDs are preferable to the corner and axial points included in CCDs. One reason might be that for the selected test functions, which model different physical processes, the non-linearity is better captured if points are distributed on the edges of the design space rather than on the axes built from the center of the design space. The predictive performance obtained using the BBD makes such a design comparable with other experimental designs that have twice the number of levels, especially in the heteroscedastic case, which is rather surprising.

Another finding is that, as expected, the addition of replicates provides a valuable contribution only as the noise increases, especially in the heteroscedastic case. However, in general, there is no proof that replicated designs should be adopted over non-replicated designs. Apparently, the exploration of less unique input configurations in replicated designs hinders the ability of the predictive models to appropriately learn the behavior of the underlying test functions, and this effect is more critical than the possibility of separating noise from signal that is provided by the presence of replicates. From a practical standpoint, it appears that replicated designs should only be chosen if the underlying phenomenon presents severe heteroscedasticity.  Again, the designs from the MAXPRO\_dis family  outperform both the competitors (D\_opt\_50\%repl and I\_opt\_50\%repl), although sometimes only marginally.

\subsection{Discussion: predictive models}
\label{S:5.2}
The unprocessed results shown in Figure \ref{bplot_piston} already convey the idea that the choice of model for data analysis is critical to the value of final predictions: given an experimental design, a \enquote{bad model} can easily provide a test error which is twice as large as the one achieved with a \enquote{good model}.

Considering the results of the first rank, a very strong, clear pattern emerges indicating that the GP model is consistently better than all its competitors for all the test functions, especially in low and intermediate homoscedastic noise settings (Fig. \ref{rankMOD_allnoise_hom}). The situation is more uncertain if we have to decide what algorithms follow in the rank.

For moderate and intermediate levels of heteroscedasticity, the GP is confirmed as the best or one of the best algorithms, but as the inhomogeneity in noise increases, some algorithms, namely SVM and RF, significantly outperform the GP (Fig. \ref{rankMOD_allnoise_het}).

The results from the second application of the ranking procedure (Tab. \ref{tab_rank_MODEL_hom}) corroborate our first impressions; in the homoscedastic case, the GP is ranked first for all the noise levels considered. The LM, SVM and ANN\_sh follow, while the RF, ANN\_dp and aml behave poorly in all the analyzed scenarios (Tab. \ref{tab_rank_MODEL_hom}).

A similar situation is shown in the final results for the heteroscedastic setting (Tab. \ref{tab_rank_MODEL_het}), but this time the SVM performs slightly better than the GP. Moreover, as the heteroscedasticity becomes severe, both the SVM and RF  outperform the GP model. Behind the GP and SVM, the LM, RF, and ANN\_sh achieve acceptable results, while, as for the homoscedastic setting, the aml and ANN\_dp perform poorly.

A general comment that can be made is that, within the scope of the present study, the GP appears to be the preferable choice for data analysis. This justifies the widespread adoption of the model for the analysis of computer experiments. However, in this case the algorithm proves its merits also when fitting designs with a moderate number of factor levels, mimicking the situation of physical experimentation. Nonetheless, attention should be paid to the training data because in the case of severe heteroscedastic noise, other options appear superior.

The SVM and LM also accomplish acceptable results, with the SVM the best performer in the heteroscedastic setting and, as expected, the LM performing well only in the homoscedastic case. Another finding worthy of note is the different performances of RF for homogeneous and inhomogeneous noise, as the algorithm increases its performance as the impact of heteroscedasticity increases, becoming one of the best options.

The results for the ANN algorithms, that we recall being by far the most adopted ML algorithms for analysis of DOE data \cite{arbo_peg21SLR}, are also interesting. The ANN\_sh is never among the best performers, neither in the homoscedastic nor heteroscedastic setting, while the ANN\_dp is the overall poorest performer among the models considered. The aml also performs badly, however it should be pointed out that both the algorithms contained in the aml platform and ANN\_dp are tools that take advantage of big data, while in this study we only consider small data for training the algorithms.

Finally, we remind the reader that in this study we only considered the prediction error as a valuable metric to rank the models, while other indicators, such as the possibility of providing a quantification of prediction uncertainty and model interpretability are crucial aspects that should also be considered \cite{peg_JAS, arbo_peg21SLR}. To this end, we mention that GPs natively provide a quantification of uncertainty and are recognized as being fairly interpretable in the ML literature \cite{GP4ML2006, liu2020gaussian}.

\section{Conclusions}
\label{S:6}
In this paper we introduce a procedure and a simulation study to provide an empirical assessment of the most promising experimental designs and predictive models to be used jointly for data collection and analysis. The focus is on physical experimentation therefore we mainly consider test functions that emulate physical processes, use small data and favor designs with a moderate number of factor levels. Several designs, predictive models, test functions and noise settings are considered, thus the study covers circumstances which are encountered in practical industrial applications, making the analysis highly relevant for practitioners.

The results show the significant potential of designs from the MAXPRO family \cite{joseph2015maximum, joseph2020designing}, especially when the factors are restricted to the discrete numeric case, reducing the number of levels to fit the budget of the experimenter. Overall, these designs outperform their competitors from the literature even when the same number of factor levels is considered, meaning that the MAXPRO criterion can maximize the ability of the predictive models to capture the non-linearity in the underlying phenomenon. Other good designs are the I-optimals, as they provide only marginally worse performances than the MAXPRO discrete numeric designs in both the homoscedastic and heteroscedastic situations. Considering ML models, the GP appears to be the overall best performer, but other methods provide better predictions in the case of severe heteroscedasticity.

Nevertheless, it should be remembered that the choice of experimental design is more critical than the choice of model for data analysis because the first must be chosen before the project starts, while the second can be tuned after the experimental data has been collected. Thus, several methods can be compared to find the best performer for the specific set of data.

\section{Acknowledgments}
The authors gratefully thank Fondazione Cariparo for partially supporting this research.

\begin{appendices}

\section{Appendix A}
\label{appA}
\addtodelayedfloat{figure}{\renewcommand{\thefigure}{A\arabic{figure}}%
  \setcounter{figure}{0}}%
In this section we report the results of the first rank for the experimental designs.

\begin{figure}[htbp]
\centering\includegraphics[width=1\linewidth]{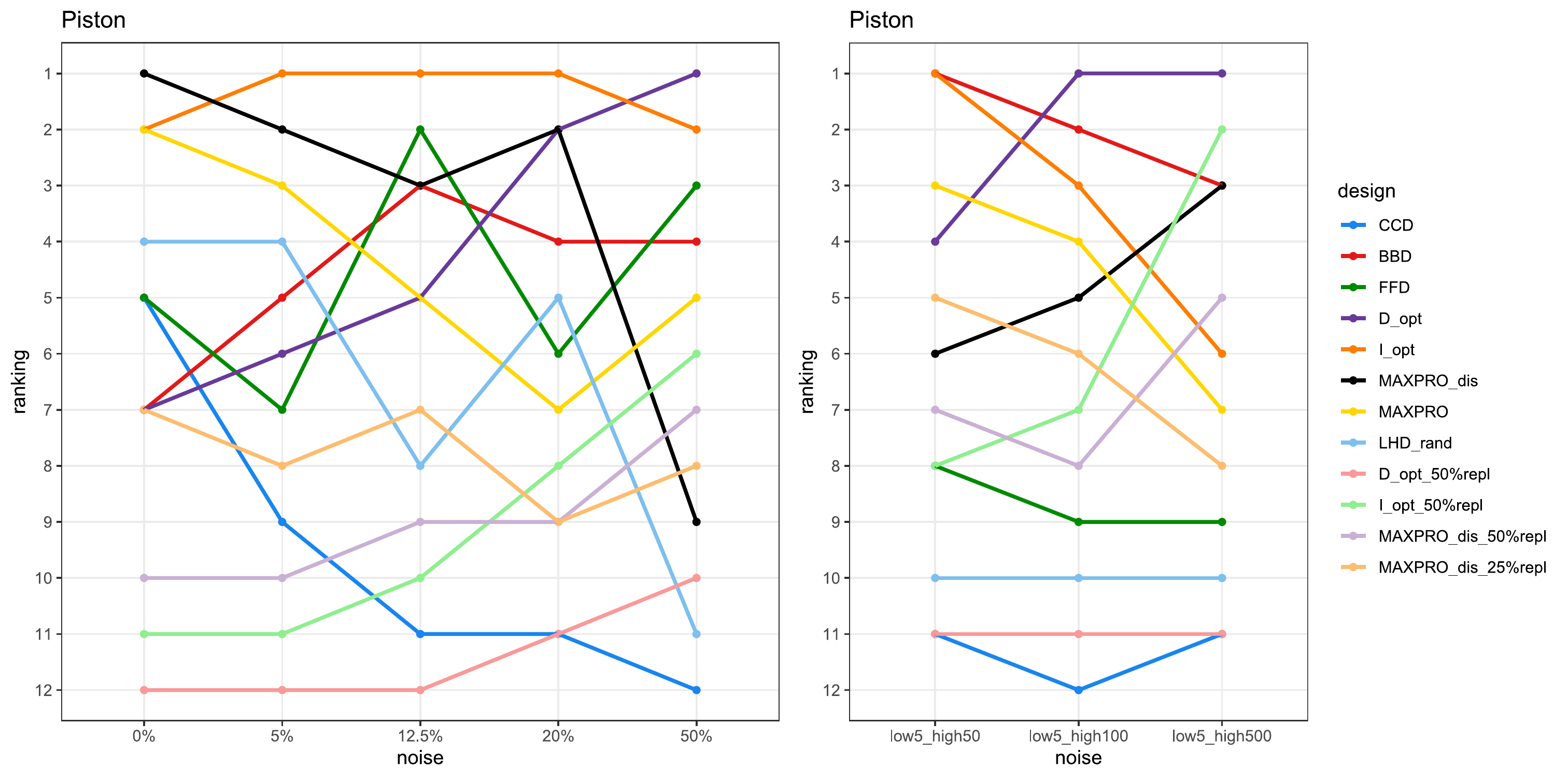}
\caption{Summary of the ranks of the experimental designs for the function \enquote{Piston}. The homoscedastic cases are reported on the left and the heteroscedastic settings are reported on the right.}
\label{1st_rank_piston}
\end{figure}

\begin{figure}[htbp]
\centering\includegraphics[width=.75\linewidth]{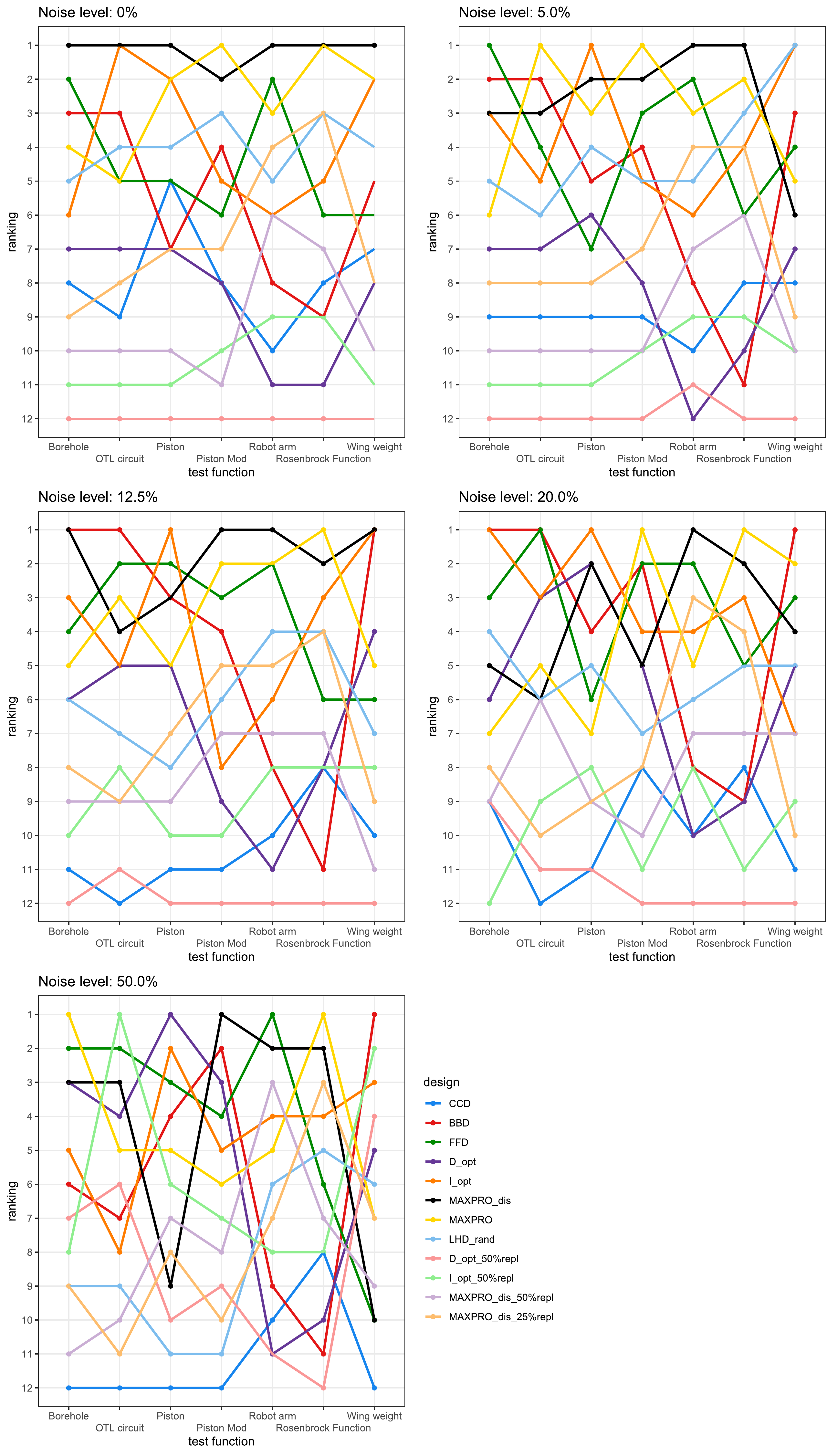}
\caption{Summary of the ranks of the experimental designs for each test function and all  homoscedastic noise settings.}
\label{rankDES_allnoise_hom}
\end{figure}

\begin{figure}[htbp]
\centering\includegraphics[width=1.05\linewidth]{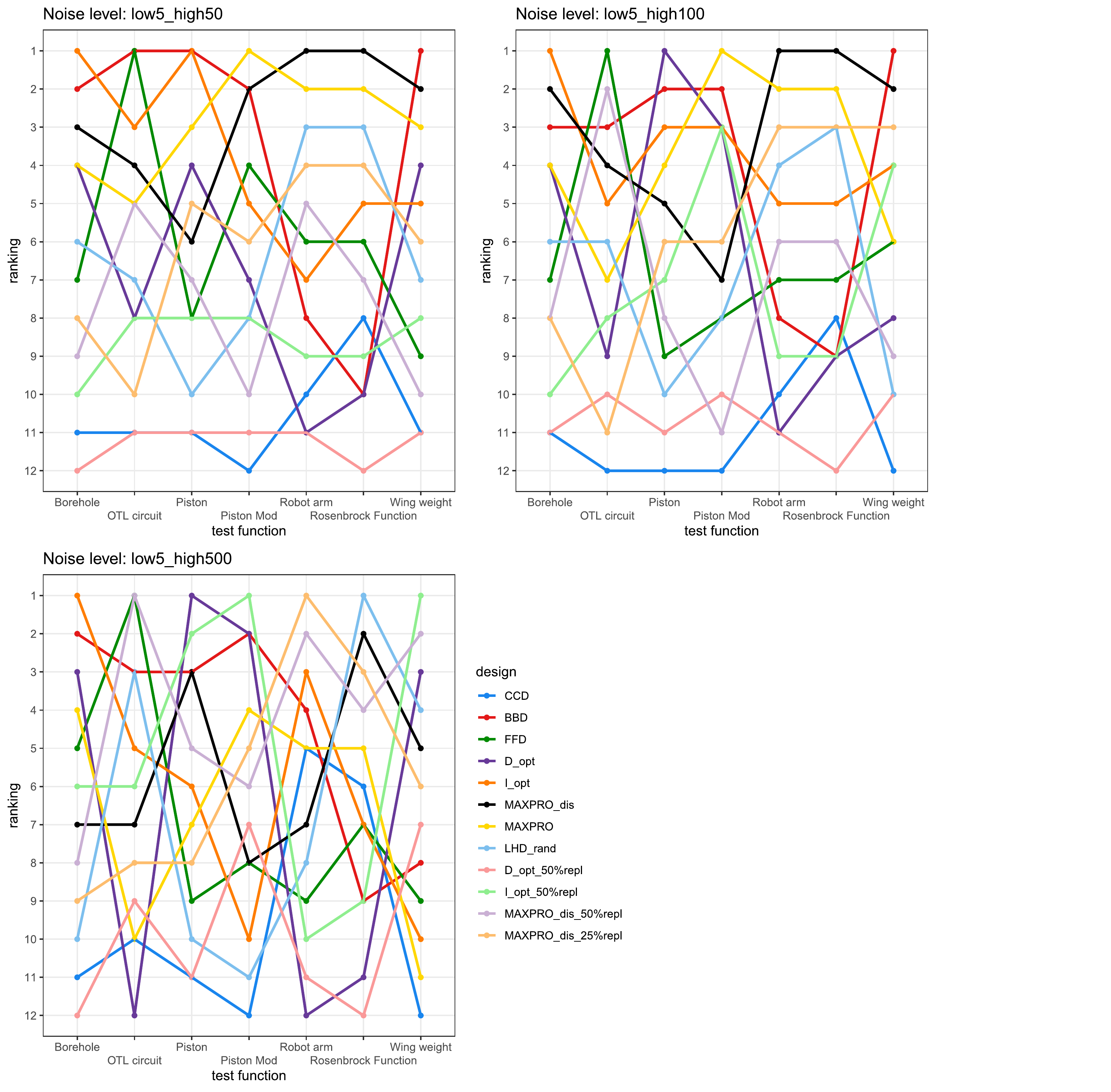}
\caption{Summary of the ranks of the experimental designs for each test function and all  heteroscedastic noise settings.}
\label{rankDES_allnoise_het}
\end{figure}

\section{Appendix B}
\label{appB}
\addtodelayedfloat{figure}{\renewcommand{\thefigure}{B\arabic{figure}}%
  \setcounter{figure}{0}}%
In this section we report the results of the first rank for the predictive models.

\begin{figure}[htbp]
\centering\includegraphics[width=.95\linewidth]{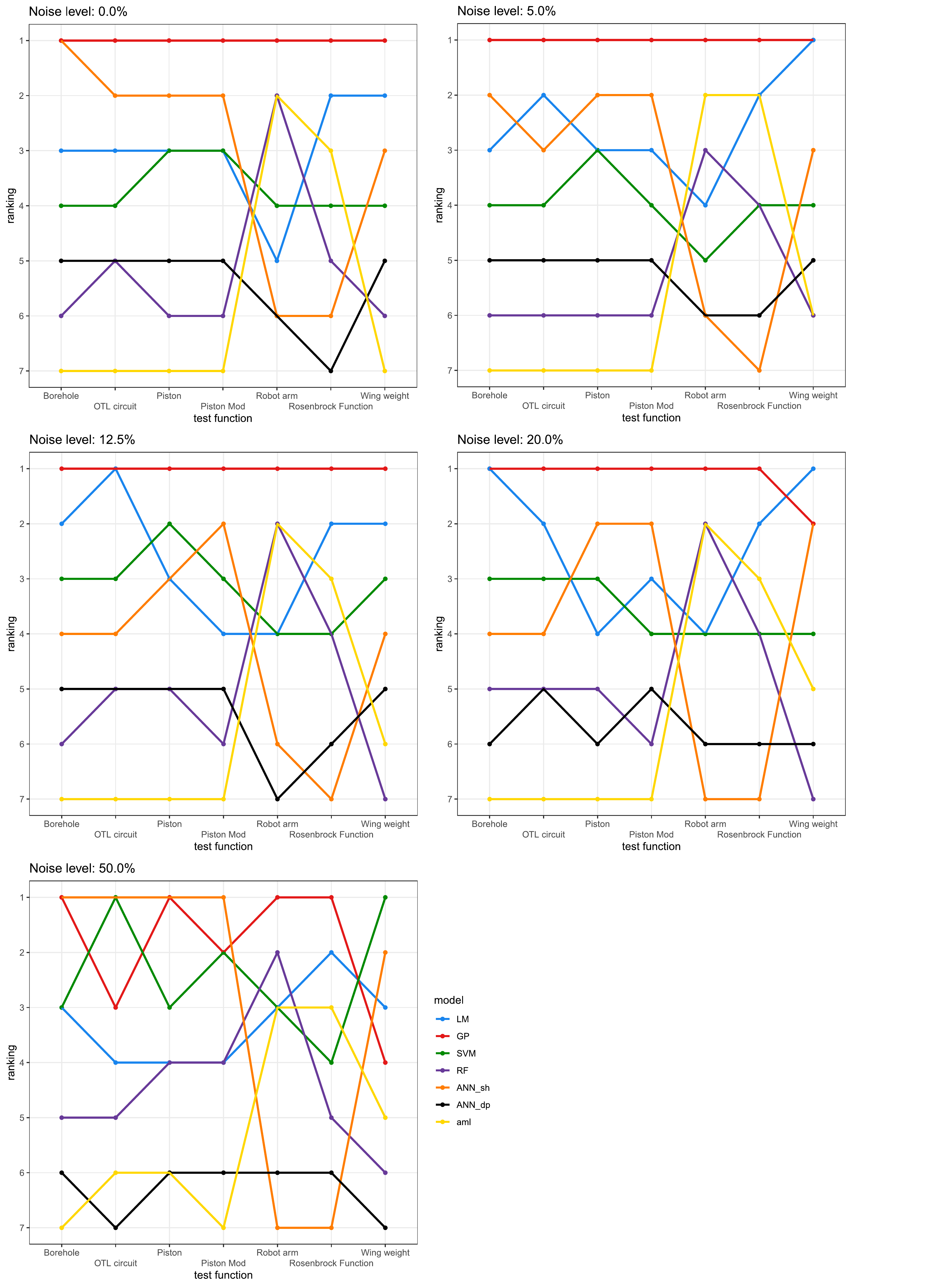}
\caption{Summary of the ranks of the predictive models for each test function and all  homoscedastic noise settings.}
\label{rankMOD_allnoise_hom}
\end{figure}

\begin{figure}[htbp]
\centering\includegraphics[width=1\linewidth]{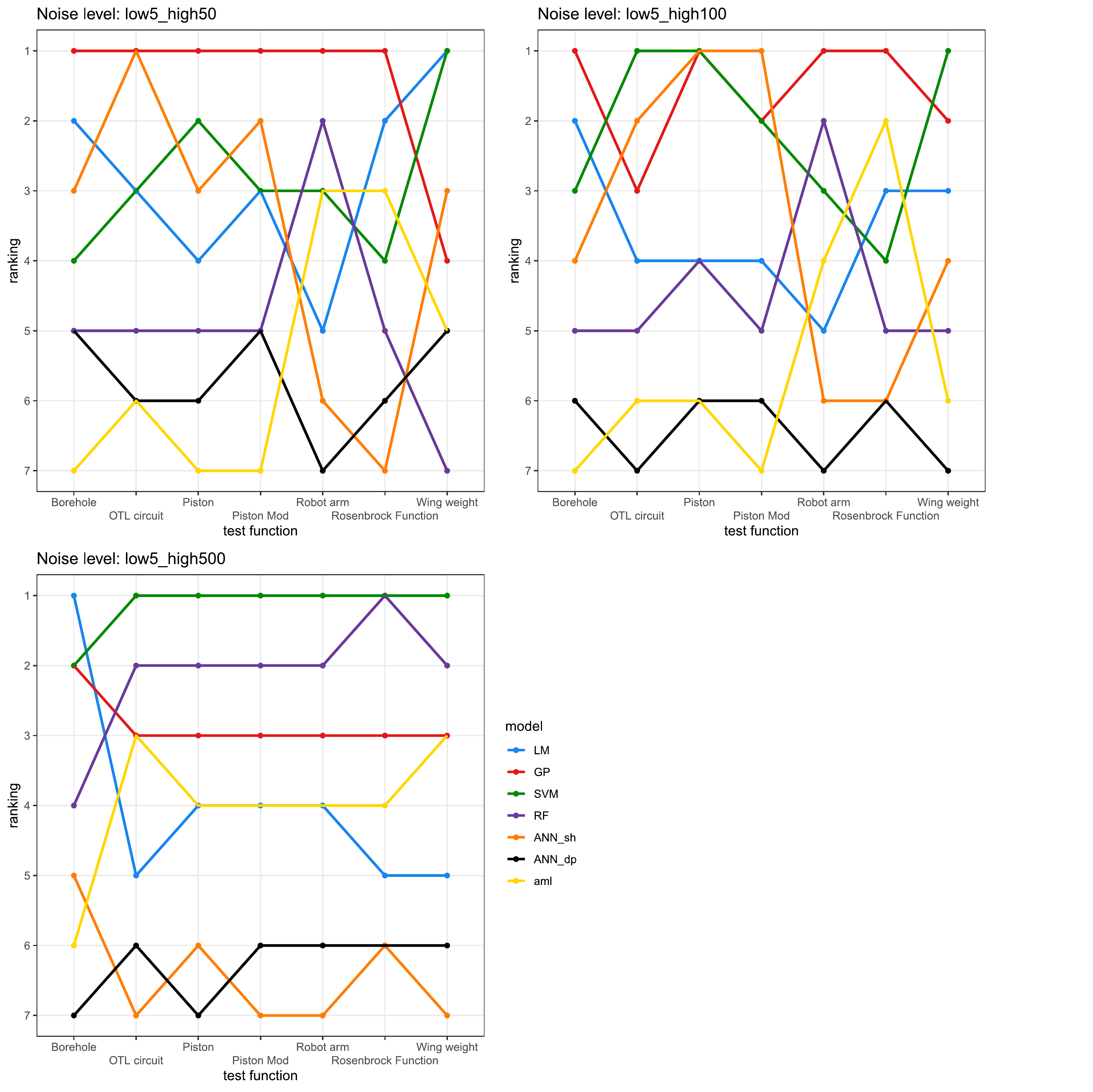}
\caption{Summary of the ranks of the predictive models for each test function and all  heteroscedastic noise settings.}
\label{rankMOD_allnoise_het}
\end{figure}

\end{appendices}

\begin{figure}[htpb]
    \centering
   \includegraphics[width=1\textwidth, page=1]{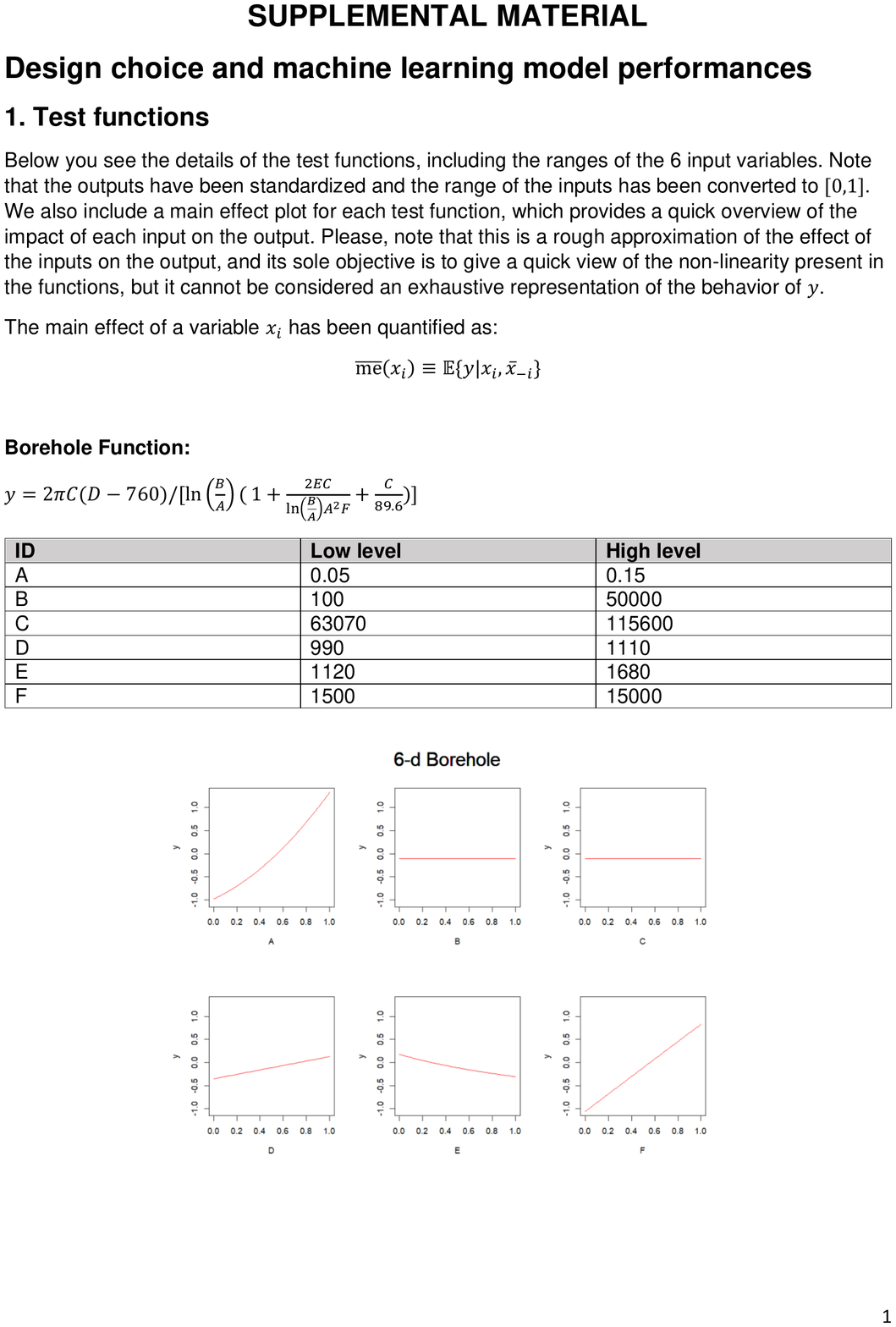}
\end{figure}
\begin{figure}[htpb]
    \centering
   \includegraphics[width=1\textwidth, page=2]{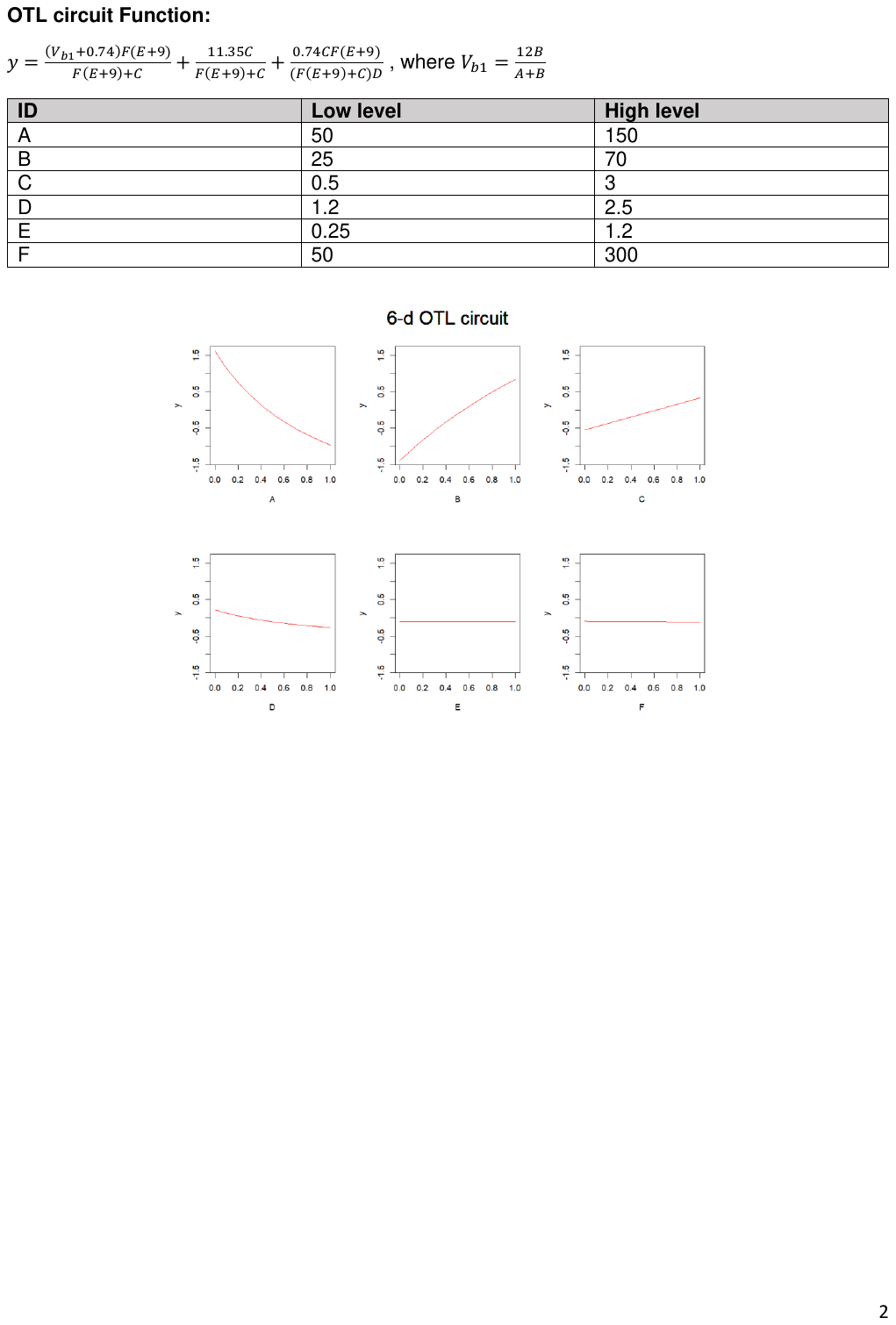}
\end{figure}

\begin{figure}[htpb]
    \centering
   \includegraphics[width=1\textwidth, page=3]{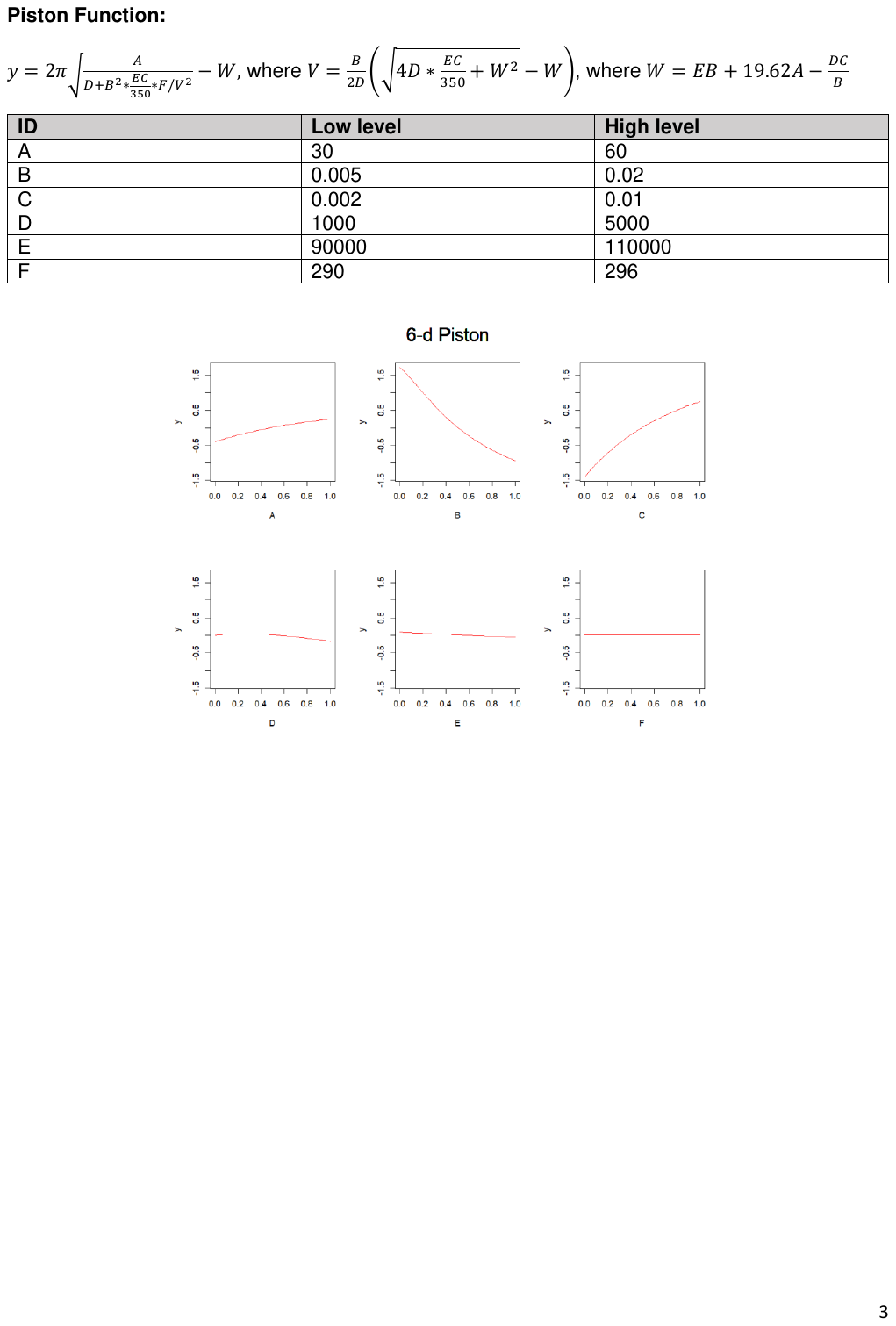}
\end{figure}
\begin{figure}[htpb]
    \centering
   \includegraphics[width=1\textwidth, page=4]{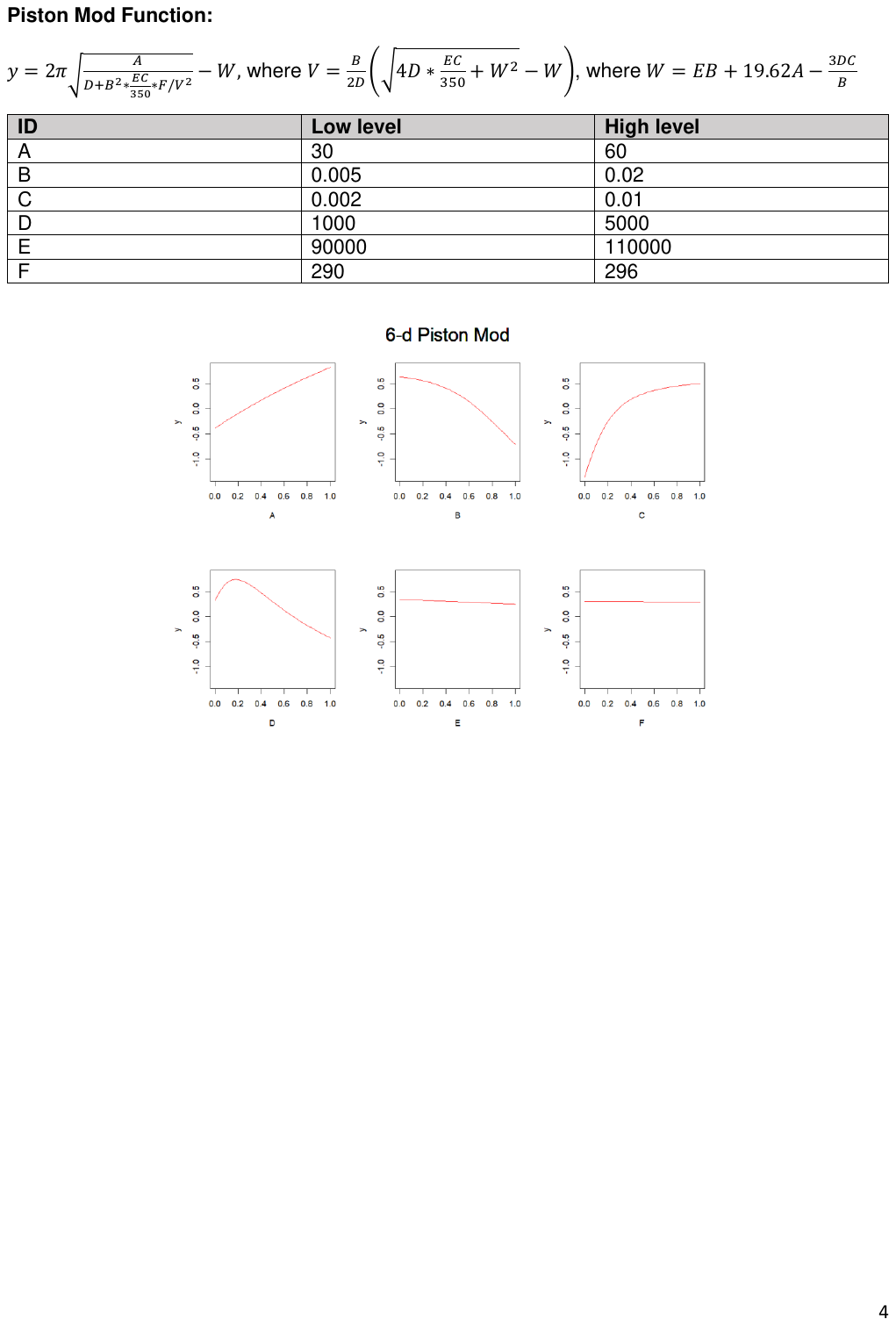}
\end{figure}
\begin{figure}[htpb]
    \centering
   \includegraphics[width=1\textwidth, page=5]{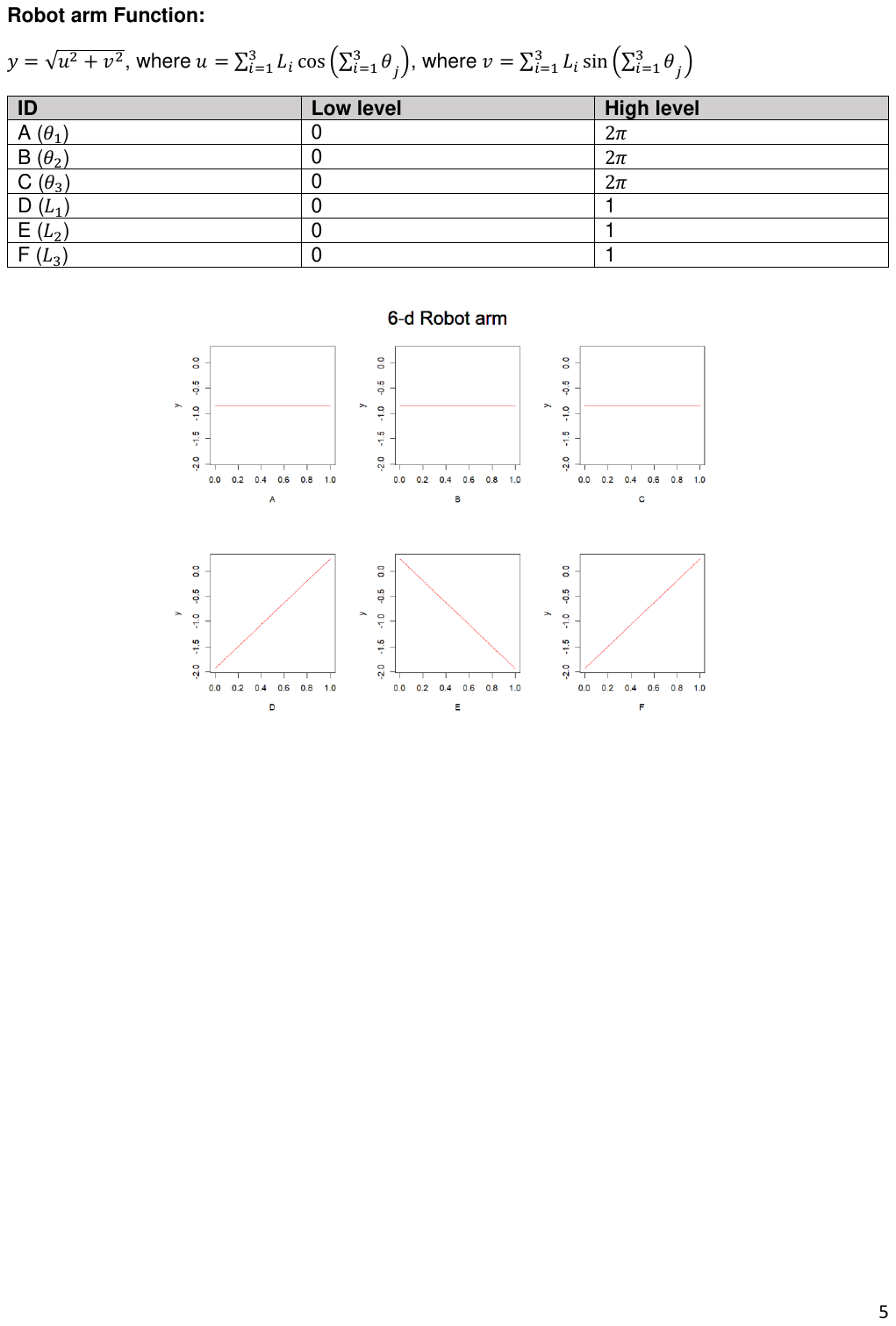}
\end{figure}
\begin{figure}[htpb]
    \centering
   \includegraphics[width=1\textwidth, page=6]{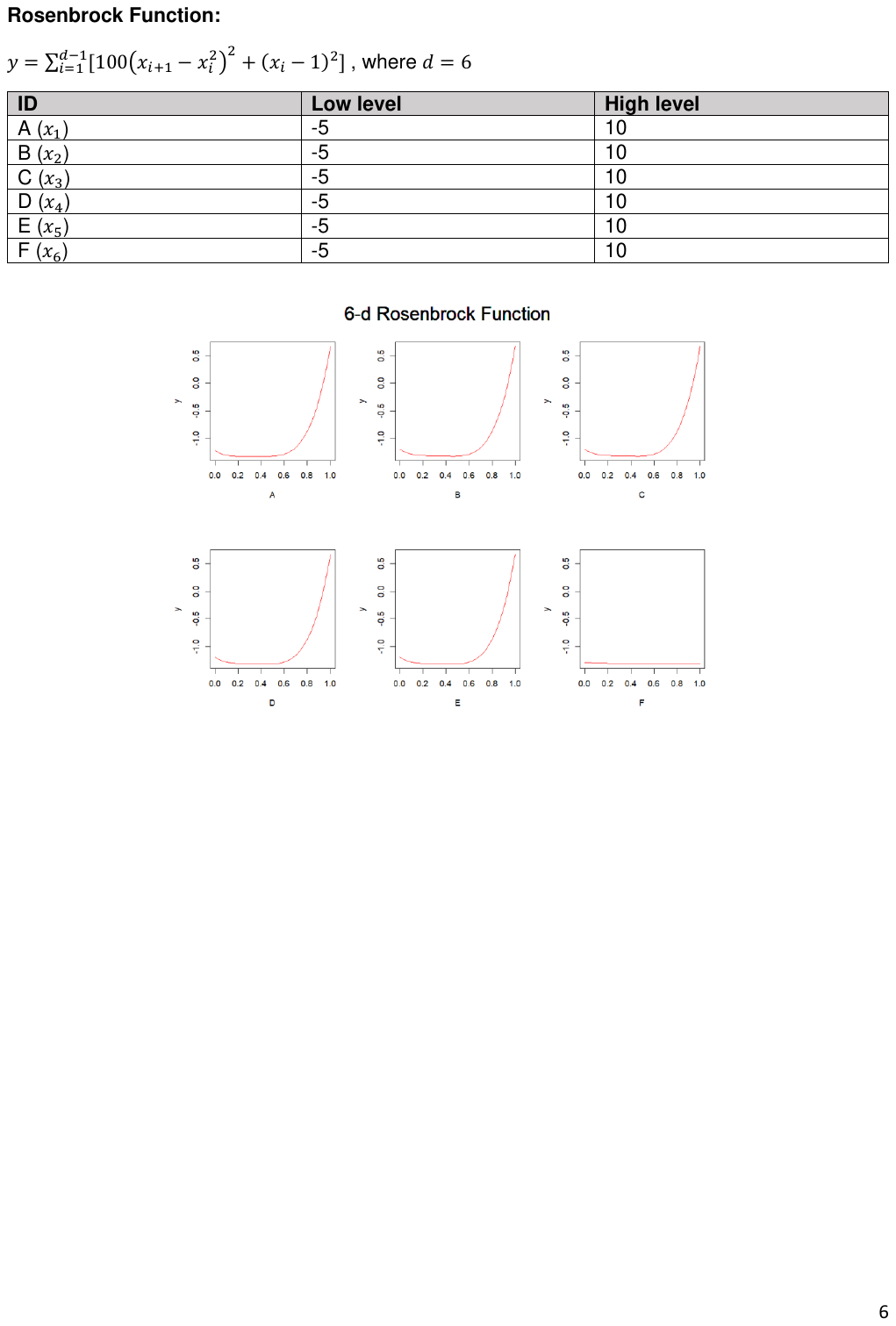}
\end{figure}

\begin{figure}[htpb]
    \centering
   \includegraphics[width=1\textwidth, page=7]{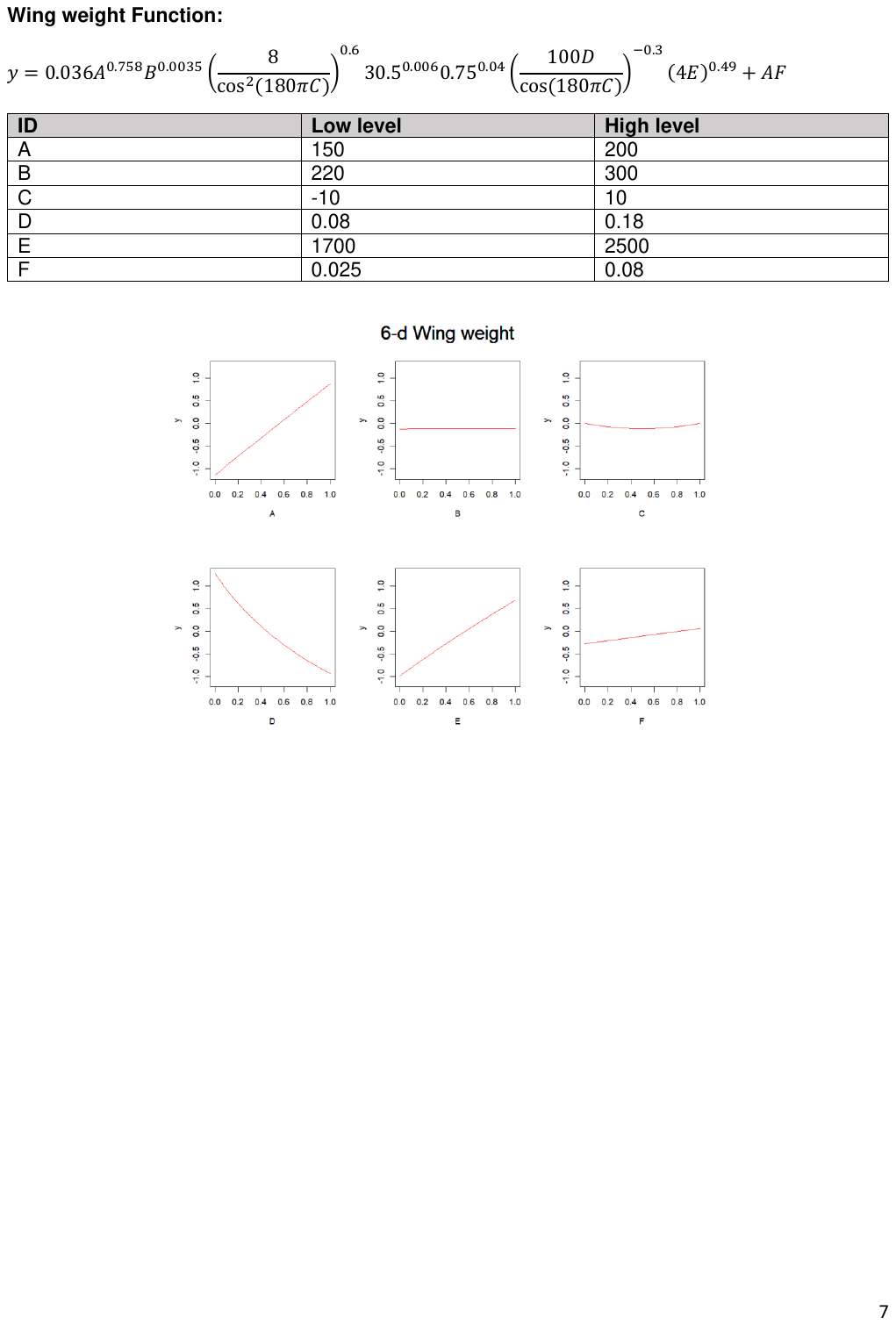}
\end{figure}


\begin{thebibliography}{58}
\expandafter\ifx\csname natexlab\endcsname\relax\def\natexlab#1{#1}\fi
\providecommand{\url}[1]{\texttt{#1}}
\providecommand{\href}[2]{#2}
\providecommand{\path}[1]{#1}
\providecommand{\DOIprefix}{doi:}
\providecommand{\ArXivprefix}{arXiv:}
\providecommand{\URLprefix}{URL: }
\providecommand{\Pubmedprefix}{pmid:}
\providecommand{\doi}[1]{\href{http://dx.doi.org/#1}{\path{#1}}}
\providecommand{\Pubmed}[1]{\href{pmid:#1}{\path{#1}}}
\providecommand{\bibinfo}[2]{#2}
\ifx\xfnm\relax \def\xfnm[#1]{\unskip,\space#1}\fi
\bibitem[{Myers et~al.(2004)Myers, Montgomery, Vining, Borror, and
  Kowalski}]{MyersMontgomery2004}
\bibinfo{author}{R.~H. Myers}, \bibinfo{author}{D.~C. Montgomery},
  \bibinfo{author}{G.~G. Vining}, \bibinfo{author}{C.~M. Borror},
  \bibinfo{author}{S.~M. Kowalski},
\newblock \bibinfo{title}{Response surface methodology: A retrospective and
  literature survey},
\newblock \bibinfo{journal}{Journal of Quality Technology} \bibinfo{volume}{36}
  (\bibinfo{year}{2004}) \bibinfo{pages}{53--77}.
  \DOIprefix\doi{10.1080/00224065.2004.11980252}.
\bibitem[{Arboretti et~al.(2021)Arboretti, Ceccato, Pegoraro, and
  Salmaso}]{arbo_peg21SLR}
\bibinfo{author}{R.~Arboretti}, \bibinfo{author}{R.~Ceccato},
  \bibinfo{author}{L.~Pegoraro}, \bibinfo{author}{L.~Salmaso},
\newblock \bibinfo{title}{Design of experiments and machine learning for
  product innovation: a systematic literature review},
\newblock \bibinfo{journal}{Quality and Reliability Engineering International}
  (\bibinfo{year}{2021}) \bibinfo{pages}{1--26}.
  \DOIprefix\doi{10.1002/qre.3025}.
\bibitem[{Box and Wilson(1951)}]{CCD}
\bibinfo{author}{G.~E.~P. Box}, \bibinfo{author}{K.~B. Wilson},
\newblock \bibinfo{title}{On the experimental attainment of optimum
  conditions},
\newblock \bibinfo{journal}{Journal of the Royal Statistical Society. Series B
  (Methodological)} \bibinfo{volume}{13} (\bibinfo{year}{1951})
  \bibinfo{pages}{1--45}. \URLprefix \url{http://www.jstor.org/stable/2983966}.
\bibitem[{Box and Behnken(1960)}]{BBD}
\bibinfo{author}{G.~E. Box}, \bibinfo{author}{D.~W. Behnken},
\newblock \bibinfo{title}{Some new three level designs for the study of
  quantitative variables},
\newblock \bibinfo{journal}{Technometrics} \bibinfo{volume}{2}
  (\bibinfo{year}{1960}) \bibinfo{pages}{455--475}.
\bibitem[{Taguchi and Wu(1979)}]{taguchi1979introduction}
\bibinfo{author}{G.~Taguchi}, \bibinfo{author}{Y.~Wu},
  \bibinfo{title}{Introduction to off-line quality control},
  \bibinfo{publisher}{Central Japan Quality Control Assoc.},
  \bibinfo{year}{1979}.
\bibitem[{Taguchi(1986)}]{taguchi1986introduction}
\bibinfo{author}{G.~Taguchi}, \bibinfo{title}{Introduction to quality
  engineering: designing quality into products and processes},
  \bibinfo{type}{Technical Report}, \bibinfo{year}{1986}.
\bibitem[{Taguchi(1987)}]{taguchi1987system}
\bibinfo{author}{G.~Taguchi}, \bibinfo{title}{System of experimental design;
  engineering methods to optimize quality and minimize costs},
  \bibinfo{type}{Technical Report}, \bibinfo{year}{1987}.
\bibitem[{Joseph(2016)}]{Joseph2016}
\bibinfo{author}{V.~R. Joseph},
\newblock \bibinfo{title}{Space-filling designs for computer experiments: A
  review},
\newblock \bibinfo{journal}{Quality Engineering} \bibinfo{volume}{28}
  (\bibinfo{year}{2016}) \bibinfo{pages}{28--35}.
  \DOIprefix\doi{10.1080/08982112.2015.1100447}.
\bibitem[{Box(1999)}]{box1999statistics}
\bibinfo{author}{G.~E. Box},
\newblock \bibinfo{title}{Statistics as a catalyst to learning by scientific
  method part ii—a discussion},
\newblock \bibinfo{journal}{Journal of Quality Technology} \bibinfo{volume}{31}
  (\bibinfo{year}{1999}) \bibinfo{pages}{16--29}.
\bibitem[{Vining(2011)}]{vining2011}
\bibinfo{author}{G.~Vining},
\newblock \bibinfo{title}{Technical advice: Design of experiments, response
  surface methodology, and sequential experimentation},
\newblock \bibinfo{journal}{Quality Engineering} \bibinfo{volume}{23}
  (\bibinfo{year}{2011}) \bibinfo{pages}{217--220}.
  \DOIprefix\doi{10.1080/15226514.2011.555280}.
\bibitem[{Jensen(2018)}]{jensen2018open}
\bibinfo{author}{W.~A. Jensen},
\newblock \bibinfo{title}{Open problems and issues in optimal design},
\newblock \bibinfo{journal}{Quality Engineering} \bibinfo{volume}{30}
  (\bibinfo{year}{2018}) \bibinfo{pages}{583--593}.
\bibitem[{Cortes et~al.(2018)Cortes, Simpson, and Parker}]{cortes2018response}
\bibinfo{author}{L.~A. Cortes}, \bibinfo{author}{J.~R. Simpson},
  \bibinfo{author}{P.~A. Parker},
\newblock \bibinfo{title}{Response surface split-plot designs: A literature
  review},
\newblock \bibinfo{journal}{Quality and Reliability Engineering International}
  \bibinfo{volume}{34} (\bibinfo{year}{2018}) \bibinfo{pages}{1374--1389}.
\bibitem[{Jones and Nachtsheim(2011)}]{jones2011class}
\bibinfo{author}{B.~Jones}, \bibinfo{author}{C.~J. Nachtsheim},
\newblock \bibinfo{title}{A class of three-level designs for definitive
  screening in the presence of second-order effects},
\newblock \bibinfo{journal}{Journal of Quality Technology} \bibinfo{volume}{43}
  (\bibinfo{year}{2011}) \bibinfo{pages}{1--15}.
\bibitem[{Chen et~al.(2016)Chen, Loeppky, Sacks, Welch
  et~al.}]{chen2016analysis}
\bibinfo{author}{H.~Chen}, \bibinfo{author}{J.~L. Loeppky},
  \bibinfo{author}{J.~Sacks}, \bibinfo{author}{W.~J. Welch}, et~al.,
\newblock \bibinfo{title}{Analysis methods for computer experiments: How to
  assess and what counts?},
\newblock \bibinfo{journal}{Statistical science} \bibinfo{volume}{31}
  (\bibinfo{year}{2016}) \bibinfo{pages}{40--60}.
\bibitem[{Ilzarbe et~al.(2008)Ilzarbe, {\'A}lvarez, Viles, and
  Tanco}]{ilzarbe2008practical}
\bibinfo{author}{L.~Ilzarbe}, \bibinfo{author}{M.~J. {\'A}lvarez},
  \bibinfo{author}{E.~Viles}, \bibinfo{author}{M.~Tanco},
\newblock \bibinfo{title}{Practical applications of design of experiments in
  the field of engineering: a bibliographical review},
\newblock \bibinfo{journal}{Quality and Reliability Engineering International}
  \bibinfo{volume}{24} (\bibinfo{year}{2008}) \bibinfo{pages}{417--428}.
\bibitem[{Morris and Mitchell(1995)}]{morris1995exploratory}
\bibinfo{author}{M.~D. Morris}, \bibinfo{author}{T.~J. Mitchell},
\newblock \bibinfo{title}{Exploratory designs for computational experiments},
\newblock \bibinfo{journal}{Journal of statistical planning and inference}
  \bibinfo{volume}{43} (\bibinfo{year}{1995}) \bibinfo{pages}{381--402}.
\bibitem[{Inc.(2019)}]{JMP}
\bibinfo{author}{S.~I. Inc.},
  \bibinfo{title}{JMP\textsuperscript{\tiny\textregistered} 14 Design of
  Experiments Guide}, \bibinfo{publisher}{SAS Institute Inc.},
  \bibinfo{year}{2019}.
\bibitem[{Zahran et~al.(2003)Zahran, Anderson-Cook, and Myers}]{FDS_plot}
\bibinfo{author}{A.~Zahran}, \bibinfo{author}{C.~M. Anderson-Cook},
  \bibinfo{author}{R.~H. Myers},
\newblock \bibinfo{title}{Fraction of design space to assess prediction
  capability of response surface designs},
\newblock \bibinfo{journal}{Journal of Quality Technology} \bibinfo{volume}{35}
  (\bibinfo{year}{2003}) \bibinfo{pages}{377--386}. \URLprefix
  \url{https://doi.org/10.1080/00224065.2003.11980235}.
  \DOIprefix\doi{10.1080/00224065.2003.11980235}.
  \href{http://arxiv.org/abs/https://doi.org/10.1080/00224065.2003.11980235}{{\tt
  arXiv:https://doi.org/10.1080/00224065.2003.11980235}}.
\bibitem[{Rodriguez et~al.(2010)Rodriguez, Jones, Borror, and
  Montgomery}]{rodriguez2010generating}
\bibinfo{author}{M.~Rodriguez}, \bibinfo{author}{B.~Jones},
  \bibinfo{author}{C.~M. Borror}, \bibinfo{author}{D.~C. Montgomery},
\newblock \bibinfo{title}{Generating and assessing exact g-optimal designs},
\newblock \bibinfo{journal}{Journal of quality technology} \bibinfo{volume}{42}
  (\bibinfo{year}{2010}) \bibinfo{pages}{3--20}.
\bibitem[{Goos et~al.(2016)Goos, Jones, and Syafitri}]{goos2016optimal}
\bibinfo{author}{P.~Goos}, \bibinfo{author}{B.~Jones},
  \bibinfo{author}{U.~Syafitri},
\newblock \bibinfo{title}{I-optimal design of mixture experiments},
\newblock \bibinfo{journal}{Journal of the American Statistical Association}
  \bibinfo{volume}{111} (\bibinfo{year}{2016}) \bibinfo{pages}{899--911}.
\bibitem[{Jones et~al.(2021)Jones, Allen-Moyer, and Goos}]{jones2020optimal}
\bibinfo{author}{B.~Jones}, \bibinfo{author}{K.~Allen-Moyer},
  \bibinfo{author}{P.~Goos},
\newblock \bibinfo{title}{A-optimal versus d-optimal design of screening
  experiments},
\newblock \bibinfo{journal}{Journal of Quality Technology} \bibinfo{volume}{53}
  (\bibinfo{year}{2021}) \bibinfo{pages}{369--382}. \URLprefix
  \url{https://doi.org/10.1080/00224065.2020.1757391}.
  \DOIprefix\doi{10.1080/00224065.2020.1757391}.
  \href{http://arxiv.org/abs/https://doi.org/10.1080/00224065.2020.1757391}{{\tt
  arXiv:https://doi.org/10.1080/00224065.2020.1757391}}.
\bibitem[{Meyer and Nachtsheim(1995)}]{coordexchalgor95}
\bibinfo{author}{R.~K. Meyer}, \bibinfo{author}{C.~J. Nachtsheim},
\newblock \bibinfo{title}{The coordinate-exchange algorithm for constructing
  exact optimal experimental designs},
\newblock \bibinfo{journal}{Technometrics} \bibinfo{volume}{37}
  (\bibinfo{year}{1995}) \bibinfo{pages}{60--69}. \URLprefix
  \url{https://www.tandfonline.com/doi/abs/10.1080/00401706.1995.10485889}.
  \DOIprefix\doi{10.1080/00401706.1995.10485889}.
  \href{http://arxiv.org/abs/https://www.tandfonline.com/doi/pdf/10.1080/00401706.1995.10485889}{{\tt
  arXiv:https://www.tandfonline.com/doi/pdf/10.1080/00401706.1995.10485889}}.
\bibitem[{Montgomery(2017)}]{montgomery2017design}
\bibinfo{author}{D.~C. Montgomery}, \bibinfo{title}{Design and analysis of
  experiments}, \bibinfo{publisher}{John wiley \& sons}, \bibinfo{year}{2017}.
\bibitem[{Ankenman et~al.(2010)Ankenman, Nelson, and Staum}]{anken2010}
\bibinfo{author}{B.~Ankenman}, \bibinfo{author}{B.~L. Nelson},
  \bibinfo{author}{J.~Staum},
\newblock \bibinfo{title}{Stochastic kriging for simulation metamodeling},
\newblock \bibinfo{journal}{Operations Research} \bibinfo{volume}{58}
  (\bibinfo{year}{2010}) \bibinfo{pages}{371--382}. \URLprefix
  \url{http://www.jstor.org/stable/40605923}.
\bibitem[{Boukouvalas et~al.(2014)Boukouvalas, Cornford, and
  Stehlík}]{BOUKOUVALAS20141088}
\bibinfo{author}{A.~Boukouvalas}, \bibinfo{author}{D.~Cornford},
  \bibinfo{author}{M.~Stehlík},
\newblock \bibinfo{title}{Optimal design for correlated processes with
  input-dependent noise},
\newblock \bibinfo{journal}{Computational Statistics \& Data Analysis}
  \bibinfo{volume}{71} (\bibinfo{year}{2014}) \bibinfo{pages}{1088--1102}.
  \URLprefix
  \url{https://www.sciencedirect.com/science/article/pii/S0167947313003484}.
  \DOIprefix\doi{https://doi.org/10.1016/j.csda.2013.09.024}.
\bibitem[{Jalali et~al.(2017)Jalali, Van~Nieuwenhuyse, and
  Picheny}]{jalali2017comparison}
\bibinfo{author}{H.~Jalali}, \bibinfo{author}{I.~Van~Nieuwenhuyse},
  \bibinfo{author}{V.~Picheny},
\newblock \bibinfo{title}{Comparison of kriging-based algorithms for simulation
  optimization with heterogeneous noise},
\newblock \bibinfo{journal}{European Journal of Operational Research}
  \bibinfo{volume}{261} (\bibinfo{year}{2017}) \bibinfo{pages}{279--301}.
\bibitem[{Binois et~al.(2019)Binois, Huang, Gramacy, and
  Ludkovski}]{binois2019replication}
\bibinfo{author}{M.~Binois}, \bibinfo{author}{J.~Huang}, \bibinfo{author}{R.~B.
  Gramacy}, \bibinfo{author}{M.~Ludkovski},
\newblock \bibinfo{title}{Replication or exploration? sequential design for
  stochastic simulation experiments},
\newblock \bibinfo{journal}{Technometrics} \bibinfo{volume}{61}
  (\bibinfo{year}{2019}) \bibinfo{pages}{7--23}.
\bibitem[{McKay et~al.(1979)McKay, Beckman, and Conover}]{McKay79}
\bibinfo{author}{M.~D. McKay}, \bibinfo{author}{R.~J. Beckman},
  \bibinfo{author}{W.~J. Conover},
\newblock \bibinfo{title}{A comparison of three methods for selecting values of
  input variables in the analysis of output from a computer code},
\newblock \bibinfo{journal}{Technometrics} \bibinfo{volume}{21}
  (\bibinfo{year}{1979}) \bibinfo{pages}{239--245}. \URLprefix
  \url{http://www.jstor.org/stable/1268522}.
\bibitem[{Stein(1987)}]{stein1987large}
\bibinfo{author}{M.~Stein},
\newblock \bibinfo{title}{Large sample properties of simulations using latin
  hypercube sampling},
\newblock \bibinfo{journal}{Technometrics} \bibinfo{volume}{29}
  (\bibinfo{year}{1987}) \bibinfo{pages}{143--151}.
\bibitem[{Joseph et~al.(2015)Joseph, Gul, and Ba}]{joseph2015maximum}
\bibinfo{author}{V.~R. Joseph}, \bibinfo{author}{E.~Gul},
  \bibinfo{author}{S.~Ba},
\newblock \bibinfo{title}{Maximum projection designs for computer experiments},
\newblock \bibinfo{journal}{Biometrika} \bibinfo{volume}{102}
  (\bibinfo{year}{2015}) \bibinfo{pages}{371--380}.
\bibitem[{Li et~al.(2006)Li, Sudarsanam, and Frey}]{li2006regularities}
\bibinfo{author}{X.~Li}, \bibinfo{author}{N.~Sudarsanam},
  \bibinfo{author}{D.~D. Frey},
\newblock \bibinfo{title}{Regularities in data from factorial experiments},
\newblock \bibinfo{journal}{Complexity} \bibinfo{volume}{11}
  (\bibinfo{year}{2006}) \bibinfo{pages}{32--45}.
\bibitem[{Joseph et~al.(2020)Joseph, Gul, and Ba}]{joseph2020designing}
\bibinfo{author}{V.~R. Joseph}, \bibinfo{author}{E.~Gul},
  \bibinfo{author}{S.~Ba},
\newblock \bibinfo{title}{Designing computer experiments with multiple types of
  factors: The maxpro approach},
\newblock \bibinfo{journal}{Journal of Quality Technology} \bibinfo{volume}{52}
  (\bibinfo{year}{2020}) \bibinfo{pages}{343--354}.
\bibitem[{Krishna et~al.(2021)Krishna, Joseph, Ba, Brenneman, and
  Myers}]{robust2021}
\bibinfo{author}{A.~Krishna}, \bibinfo{author}{V.~R. Joseph},
  \bibinfo{author}{S.~Ba}, \bibinfo{author}{W.~A. Brenneman},
  \bibinfo{author}{W.~R. Myers},
\newblock \bibinfo{title}{Robust experimental designs for model calibration},
\newblock \bibinfo{journal}{Journal of Quality Technology} \bibinfo{volume}{0}
  (\bibinfo{year}{2021}) \bibinfo{pages}{1--12}. \URLprefix
  \url{https://doi.org/10.1080/00224065.2021.1930618}.
  \DOIprefix\doi{10.1080/00224065.2021.1930618}.
  \href{http://arxiv.org/abs/https://doi.org/10.1080/00224065.2021.1930618}{{\tt
  arXiv:https://doi.org/10.1080/00224065.2021.1930618}}.
\bibitem[{Hastie et~al.(2009)Hastie, Tibshirani, and
  Friedman}]{hastie2009elements}
\bibinfo{author}{T.~Hastie}, \bibinfo{author}{R.~Tibshirani},
  \bibinfo{author}{J.~Friedman}, \bibinfo{title}{The elements of statistical
  learning: data mining, inference, and prediction},
  \bibinfo{publisher}{Springer Science \& Business Media},
  \bibinfo{year}{2009}.
\bibitem[{Gramacy(2020)}]{gramacy2020surrogates}
\bibinfo{author}{R.~B. Gramacy}, \bibinfo{title}{Surrogates: {G}aussian Process
  Modeling, Design and Optimization for the Applied Sciences},
  \bibinfo{publisher}{Chapman Hall/CRC}, \bibinfo{address}{Boca Raton,
  Florida}, \bibinfo{year}{2020}.
  \bibinfo{note}{\url{http://bobby.gramacy.com/surrogates/}}.
\bibitem[{Breiman(2001)}]{breiman2001random}
\bibinfo{author}{L.~Breiman},
\newblock \bibinfo{title}{Random forests},
\newblock \bibinfo{journal}{Machine learning} \bibinfo{volume}{45}
  (\bibinfo{year}{2001}) \bibinfo{pages}{5--32}.
\bibitem[{LeDell and Poirier(2020)}]{ledell2020h2o}
\bibinfo{author}{E.~LeDell}, \bibinfo{author}{S.~Poirier},
\newblock \bibinfo{title}{H2o automl: Scalable automatic machine learning},
\newblock in: \bibinfo{booktitle}{Proceedings of the AutoML Workshop at ICML},
  volume \bibinfo{volume}{2020}, \bibinfo{year}{2020}.
\bibitem[{James et~al.(2013)James, Witten, Hastie, and
  Tibshirani}]{james2013introduction}
\bibinfo{author}{G.~James}, \bibinfo{author}{D.~Witten},
  \bibinfo{author}{T.~Hastie}, \bibinfo{author}{R.~Tibshirani},
  \bibinfo{title}{An introduction to statistical learning}, volume
  \bibinfo{volume}{112}, \bibinfo{publisher}{Springer}, \bibinfo{year}{2013}.
\bibitem[{Candel and LeDell(2021)}]{candel2021deep}
\bibinfo{author}{A.~Candel}, \bibinfo{author}{E.~LeDell}, \bibinfo{title}{Deep
  learning with H2O}, \bibinfo{year}{2021}. \URLprefix
  \url{http://docs.h2o.ai/h2o/resources/}, \bibinfo{note}{sixth Edition}.
\bibitem[{Srivastava et~al.(2014)Srivastava, Hinton, Krizhevsky, Sutskever, and
  Salakhutdinov}]{srivastava2014dropout}
\bibinfo{author}{N.~Srivastava}, \bibinfo{author}{G.~Hinton},
  \bibinfo{author}{A.~Krizhevsky}, \bibinfo{author}{I.~Sutskever},
  \bibinfo{author}{R.~Salakhutdinov},
\newblock \bibinfo{title}{Dropout: a simple way to prevent neural networks from
  overfitting},
\newblock \bibinfo{journal}{The journal of machine learning research}
  \bibinfo{volume}{15} (\bibinfo{year}{2014}) \bibinfo{pages}{1929--1958}.
\bibitem[{Arboretti et~al.(2021)Arboretti, Ceccato, Pegoraro, Salmaso,
  Housmekerides, Spadoni, Pierangelo, Quaggia, Tveit, and Vianello}]{peg_JAS}
\bibinfo{author}{R.~Arboretti}, \bibinfo{author}{R.~Ceccato},
  \bibinfo{author}{L.~Pegoraro}, \bibinfo{author}{L.~Salmaso},
  \bibinfo{author}{C.~Housmekerides}, \bibinfo{author}{L.~Spadoni},
  \bibinfo{author}{E.~Pierangelo}, \bibinfo{author}{S.~Quaggia},
  \bibinfo{author}{C.~Tveit}, \bibinfo{author}{S.~Vianello},
\newblock \bibinfo{title}{Machine learning and design of experiments with an
  application to product innovation in the chemical industry},
\newblock \bibinfo{journal}{Journal of Applied Statistics} \bibinfo{volume}{0}
  (\bibinfo{year}{2021}) \bibinfo{pages}{1--26}. \URLprefix
  \url{https://doi.org/10.1080/02664763.2021.1907840}.
  \DOIprefix\doi{10.1080/02664763.2021.1907840}.
\bibitem[{Karatzoglou et~al.(2004)Karatzoglou, Smola, Hornik, and
  Zeileis}]{kernlab}
\bibinfo{author}{A.~Karatzoglou}, \bibinfo{author}{A.~Smola},
  \bibinfo{author}{K.~Hornik}, \bibinfo{author}{A.~Zeileis},
\newblock \bibinfo{title}{kernlab -- an {S4} package for kernel methods in
  {R}},
\newblock \bibinfo{journal}{Journal of Statistical Software}
  \bibinfo{volume}{11} (\bibinfo{year}{2004}) \bibinfo{pages}{1--20}.
  \URLprefix \url{http://www.jstatsoft.org/v11/i09/}.
\bibitem[{Rasmussen and Williams(2006)}]{GP4ML2006}
\bibinfo{author}{C.~E. Rasmussen}, \bibinfo{author}{C.~K.~I. Williams},
  \bibinfo{title}{Gaussian Processes for Machine Learning},
  \bibinfo{publisher}{The MIT Press}, \bibinfo{year}{2006}.
\bibitem[{Roustant et~al.(2012)Roustant, Ginsbourger, and
  Deville}]{roustant2012dicekriging}
\bibinfo{author}{O.~Roustant}, \bibinfo{author}{D.~Ginsbourger},
  \bibinfo{author}{Y.~Deville},
\newblock \bibinfo{title}{Dicekriging, diceoptim: Two r packages for the
  analysis of computer experiments by kriging-based metamodeling and
  optimization},
\newblock \bibinfo{journal}{Journal of Statistical Software, Articles}
  \bibinfo{volume}{51} (\bibinfo{year}{2012}) \bibinfo{pages}{1--55}.
  \URLprefix \url{https://www.jstatsoft.org/v051/i01}.
  \DOIprefix\doi{10.18637/jss.v051.i01}.
\bibitem[{Gramacy and Lee(2012)}]{gramacy2012cases}
\bibinfo{author}{R.~B. Gramacy}, \bibinfo{author}{H.~K. Lee},
\newblock \bibinfo{title}{Cases for the nugget in modeling computer
  experiments},
\newblock \bibinfo{journal}{Statistics and Computing} \bibinfo{volume}{22}
  (\bibinfo{year}{2012}) \bibinfo{pages}{713--722}.
\bibitem[{H2O(2021)}]{amlweb}
\bibinfo{author}{H2O}, \bibinfo{title}{Automl: Automatic machine learning},
  \bibinfo{year}{2021}. \URLprefix
  \url{https://docs.h2o.ai/h2o/latest-stable/h2o-docs/automl.html}.
\bibitem[{Cook(2016)}]{cook2016practical}
\bibinfo{author}{D.~Cook}, \bibinfo{title}{Practical machine learning with H2O:
  powerful, scalable techniques for deep learning and AI},
  \bibinfo{publisher}{" O'Reilly Media, Inc."}, \bibinfo{year}{2016}.
\bibitem[{Landry(2020)}]{landry2020machine}
\bibinfo{author}{M.~Landry}, \bibinfo{title}{Machine Learning with R and H2O},
  \bibinfo{year}{2020}. \URLprefix
  \url{http://h2o-release.s3.amazonaws.com/h2o/rel-zeno/2/docs-website/h2o-docs/booklets/RBooklet.pdf},
  \bibinfo{note}{seventh Edition}.
\bibitem[{Surjanovic and Bingham(2021)}]{simulationlib}
\bibinfo{author}{S.~Surjanovic}, \bibinfo{author}{D.~Bingham},
  \bibinfo{title}{Virtual library of simulation experiments: Test functions and
  datasets}, \bibinfo{year}{2021}. \URLprefix
  \url{http://www.sfu.ca/~ssurjano}.
\bibitem[{Picheny et~al.(2013)Picheny, Wagner, and
  Ginsbourger}]{picheny2013benchmark}
\bibinfo{author}{V.~Picheny}, \bibinfo{author}{T.~Wagner},
  \bibinfo{author}{D.~Ginsbourger},
\newblock \bibinfo{title}{A benchmark of kriging-based infill criteria for
  noisy optimization},
\newblock \bibinfo{journal}{Structural and Multidisciplinary Optimization}
  \bibinfo{volume}{48} (\bibinfo{year}{2013}) \bibinfo{pages}{607--626}.
\bibitem[{Pesarin and Salmaso(2010)}]{pesarin2010permutation}
\bibinfo{author}{F.~Pesarin}, \bibinfo{author}{L.~Salmaso},
  \bibinfo{title}{Permutation tests for complex data: theory, applications and
  software}, \bibinfo{publisher}{John Wiley \& Sons}, \bibinfo{year}{2010}.
\bibitem[{Gupta and Panchapakesan(2002)}]{gupta2002}
\bibinfo{author}{S.~S. Gupta}, \bibinfo{author}{S.~Panchapakesan},
  \bibinfo{title}{Multiple Decision Procedures}, \bibinfo{publisher}{Society
  for Industrial and Applied Mathematics}, \bibinfo{year}{2002}. \URLprefix
  \url{https://epubs.siam.org/doi/abs/10.1137/1.9780898719161}.
  \DOIprefix\doi{10.1137/1.9780898719161}.
  \href{http://arxiv.org/abs/https://epubs.siam.org/doi/pdf/10.1137/1.9780898719161}{{\tt
  arXiv:https://epubs.siam.org/doi/pdf/10.1137/1.9780898719161}}.
\bibitem[{Arboretti et~al.(2014)Arboretti, Bonnini, Corain, and
  Salmaso}]{arboretti2014permutation}
\bibinfo{author}{R.~Arboretti}, \bibinfo{author}{S.~Bonnini},
  \bibinfo{author}{L.~Corain}, \bibinfo{author}{L.~Salmaso},
\newblock \bibinfo{title}{A permutation approach for ranking of multivariate
  populations},
\newblock \bibinfo{journal}{Journal of Multivariate Analysis}
  \bibinfo{volume}{132} (\bibinfo{year}{2014}) \bibinfo{pages}{39--57}.
\bibitem[{Corain et~al.(2016)Corain, Arboretti, and Bonnini}]{rankbook16}
\bibinfo{author}{L.~Corain}, \bibinfo{author}{R.~Arboretti},
  \bibinfo{author}{S.~Bonnini}, \bibinfo{title}{Ranking of Multivariate
  Populations: A Permutation Approach with Applications},
  \bibinfo{publisher}{Chapman Hall/CRC}, \bibinfo{address}{Boca Raton,
  Florida}, \bibinfo{year}{2016}. \DOIprefix\doi{10.1201/b19673}.
\bibitem[{Arboretti et~al.(2021)Arboretti, Ceccato, Pegoraro, and
  Salmaso}]{arboretti2021interval}
\bibinfo{author}{R.~Arboretti}, \bibinfo{author}{R.~Ceccato},
  \bibinfo{author}{L.~Pegoraro}, \bibinfo{author}{L.~Salmaso},
\newblock \bibinfo{title}{Interval selection: A case-study-based approach},
\newblock \bibinfo{journal}{Applied Stochastic Models in Business and Industry}
   (\bibinfo{year}{2021}).
\bibitem[{Liu et~al.(2020)Liu, Ong, Shen, and Cai}]{liu2020gaussian}
\bibinfo{author}{H.~Liu}, \bibinfo{author}{Y.-S. Ong},
  \bibinfo{author}{X.~Shen}, \bibinfo{author}{J.~Cai},
\newblock \bibinfo{title}{When gaussian process meets big data: A review of
  scalable gps},
\newblock \bibinfo{journal}{IEEE transactions on neural networks and learning
  systems} \bibinfo{volume}{31} (\bibinfo{year}{2020})
  \bibinfo{pages}{4405--4423}.
\bibitem[{Groemping(2020)}]{doewrapper}
\bibinfo{author}{U.~Groemping}, \bibinfo{title}{DoE.wrapper: Wrapper Package
  for Design of Experiments Functionality}, \bibinfo{year}{2020}. \URLprefix
  \url{https://CRAN.R-project.org/package=DoE.wrapper}, \bibinfo{note}{r
  package version 0.11}.
\bibitem[{Dupuy et~al.(2015)Dupuy, Helbert, and Franco}]{dicedesign}
\bibinfo{author}{D.~Dupuy}, \bibinfo{author}{C.~Helbert},
  \bibinfo{author}{J.~Franco},
\newblock \bibinfo{title}{{DiceDesign} and {DiceEval}: Two {R} packages for
  design and analysis of computer experiments},
\newblock \bibinfo{journal}{Journal of Statistical Software}
  \bibinfo{volume}{65} (\bibinfo{year}{2015}) \bibinfo{pages}{1--38}.
  \URLprefix \url{http://www.jstatsoft.org/v65/i11/}.

\end{thebibliography}
\end{document}